\documentclass[sigconf]{acmart}
\renewcommand\footnotetextcopyrightpermission[1]{} 
\usepackage[font=small,labelfont=bf,skip=0pt]{caption}
\usepackage{enumerate}
\usepackage{amsmath}
\usepackage{amsfonts}
\usepackage{multirow}
\usepackage{tabularx}
\usepackage{mathtools}
\usepackage{balance}
\usepackage{subfigure}
\usepackage{bbm}
\usepackage{color}
\usepackage{float}
\usepackage{listings}
\usepackage{color}
\usepackage{booktabs}
\usepackage{tablefootnote}
\usepackage{url}
\usepackage{threeparttable}

\newcommand\possiblebreak{\ifhmode\unskip\space\hfil\penalty0\hfilneg\fi}

\usepackage{graphicx}
\usepackage{color}
\usepackage{adjustbox}

\usepackage{algorithmic}
\usepackage[ruled,vlined]{algorithm2e}
\SetAlFnt{\small}

\begin{document}
\sloppy

%
\title{Analyzing and Improving Fault Tolerance of Learning-Based Navigation Systems}

\author{Zishen Wan$^{1 *}$, Aqeel Anwar$^1$, Yu-Shun Hsiao$^2$, Tianyu Jia$^2$}
\author{Vijay Janapa Reddi$^2$, Arijit Raychowdhury$^1$}
\affiliation{
  \institution{$^1$Georgia Institute of Technology, Atlanta, GA \hspace{0.05in} $^2$Harvard University, Cambridge, MA}
}
\affiliation{
\normalsize \hspace{-0.1in} *Email: zishenwan@gatech.edu
}

\markboth{Journal of \LaTeX\ Class Files,~Vol.~13, No.~9, September~2014}%
{Shell \MakeLowercase{\textit{et al.}}: Bare Demo of IEEEtran.cls for Journals}
%
\begin{abstract}
Learning-based navigation systems are widely used in autonomous applications, such as robotics, unmanned vehicles and drones. Specialized hardware accelerators have been proposed for high-performance and energy-efficiency for such navigational tasks. However, transient and permanent faults are increasing in hardware systems and can catastrophically violate tasks safety. Meanwhile, traditional redundancy-based protection methods are challenging to deploy on resource-constrained edge applications. In this paper, we experimentally evaluate the resilience of navigation systems with respect to algorithms, fault models and data types from both RL training and inference. We further propose two efficient fault mitigation techniques that achieve 2$\times$ success rate and 39\% quality-of-flight improvement in learning-based navigation systems.

\end{abstract}

\maketitle




%
\section{Introduction}
\label{intro}
Autonomous navigation continues to be deployed and attains significant traction in the field of robotics, unmanned drones and autonomous vehicles at all computing scales~\cite{liu2021robotic,wan2021survey,suleiman2019navion,gao2021ielas,krishnan2021autosoc,wan2021energy}. It helps an agent avoid navigating in unknown environments and situations; and take safe actions as needed. Recently, end-to-end learning-based techniques~\cite{anwar2018navren,krishnan2019air,duisterhof2019learning} have demonstrated considerable potential in navigation along with specialized hardware accelerators~\cite{amaravati201855,krishnan2021machine}, where the agent processes raw sensor information and uses policy trained by reinforcement learning (RL) to directly produce output actions. 

While various approaches to achieve energy-efficient and accurate navigation have been extensively studied, the inherent resilience of learning-based navigation systems to hardware faults is not well understood. Hardware accelerators are susceptible to permanent and transient faults, as a consequence of manufacturing defects or single event upset. The increasing rate of faults in technology nodes can cause application failures and safety violations. 

Faults may greatly impact the safety of navigation systems in both training and inference.
In the context of training, different from conventional offline training and deployment method (i.e., train the policy on cloud and conduct inference on edge), the learning-based system usually requires real-time training and fine-tuning on edge nodes because the discrepancy between offline simulated and new unknown environments will degrade the policy's performance~\cite{anwar2020autonomous}. Faults occurred during training might impact the learning process and even negatively affect the policy convergence due to the complexity of learning algorithms and optimization procedures.
In the context of inference, different from DNN models in supervised learning, RL policy's quality depends on how effective it is in long-term sequential decision making. Faults at one stage might propagate to subsequent stages, suggesting that policies might be more vulnerable to faults than traditional DNN applications.

Traditional methods to protect compute systems from hardware faults typically adopt ECC~\cite{tawada2015bit} or replicate hardware components~\cite{matsuo2018dual,bannon2019computer}. While these approaches are effective, they bring large overhead in the hardware cost and energy which are challenging to deploy on edge devices. In addition, this level of protection imposes a high design effort at both micro-architecture and device level, which may be an overkill for learning-based navigation policies. 

In this paper, we perform an in-depth fault characterization study on the learning-based navigation system from small (grid-based navigation task) to large (drone navigation task) computing scales. The resilience against permanent and transient faults is explored in both training and inference stages. Our results demonstrate that different learning algorithms exhibit different sensitivities to faults. We disclose the relationship between fault type, bit error rate, policy convergence and exploration rate in the learning-based system. Moreover, the system resilience depends on the model topology, data type and faulty bit location. Based on our observations, we devise two cost-effective solutions to mitigate faults impact: adjustment of the ratio of exploration-to-exploitation during RL training and range-based anomaly detection during inference.

This paper, therefore, makes the following contributions:
\begin{itemize}
    \item We develop a fault injection tool-chain that emulates various types of hardware faults and enables rapid fault analysis of learning-based navigation systems.
    \item We conduct a large-scale fault injection study in both training and inference stages of learning-based systems against permanent and transient faults. We show how faults can impact policy convergence and exploration with different training algorithms and evaluate the system resilience dependent on model structure, data type and faulty bit position. To the best of our knowledge, this is the first such comprehensive study of hardware faults in RL.
    \item We present two low-overhead techniques to mitigate transient and permanent fault impacts in both RL training and inference. By dynamically adjusting exploration rate in training and detecting anomaly values in inference, the learning-based system achieves up to 2$\times$ resilience improvement.
\end{itemize}
\section{Related Work}
\label{sec:related} 
\noindent \textbf{Reliability of autonomous system.} The resilience of navigation system has been recently studied in the autonomous vehicle field. \cite{toschi2019characterizing} assesses the noise in sensor, while we focus on the hardware faults during compute. \cite{jha2019ml,hsiao2021mavfi} evaluate hardware faults on the traditional kernel-based pipeline consisting of several function modules such as perception, localization and planning. However, the reliability of the emerging end-to-end learning-based navigation system has not been adequately explored, which is the focus of our work.



\noindent \textbf{Fault characterization.} The impact of faults in neural network, a core component in the learning-based navigation system, have been studied in a wide body of recent work. \cite{reagen2018ares, mahmoud2020pytorchfi,chen2020tensorfi} build fault injection frameworks to quantify the error resilience of DNN applications. \cite{li2017understanding} evaluated the fault propagation of DNN focused on the vulnerability of different layers in an ASIC model. However, these works exclusively focus on the resiliency of neural networks to hardware faults without considering the inherent resilience for end-to-end learning-based safety-critical systems. Unlike a single neural network model in supervised learning, the resiliency of models (i.e., policies) in the learning-based system depends on the sequential decision-making process. Therefore, we take the first step to explore how hardware faults affect the training process and long-term decision-making capability of learning-based systems.

\noindent \textbf{Fault mitigation techniques.} Several general techniques have been proposed to mitigate faults impacts, e.g., DMR~\cite{matsuo2018dual}, TMR~\cite{hudson2018fault} and ECC~\cite{tawada2015bit}. Some NN-specified techniques, such as explicit redundancy and retraining, are proposed in~\cite{torres2017fault}.
These techniques can be effective for fault detection and recovery, but incur area, power or timing overhead which is challenging to deploy in resource-limited edge nodes. Instead of costly fault tolerance operations, we present application-aware fault mitigation techniques while considering the characteristics of the learning-based system and bit significance. The proposed techniques don't require redundant bits and can be applied to both RL training and inference stages.




\section{Methodology and Fault Model} 
\label{sec:meth}
\subsection{Learning-Based Navigational System}
\label{subsec:RL_methodology}

We consider the task of reinforcement learning based navigational system \cite{anwar2018navren}. The goal of such system is to learn a navigational policy by interacting with the underlying environment achieving user-defined goals. We learn these navigational policies with the help of RL where the agent constantly interacts with the environment $\mathcal{E}$ characterized by a Markov Decision Process (MDP). The MDP of the environment can be described by the tuple $\mathcal{M} = (\mathcal{S},\mathcal{A},\mathcal{P},\mathcal{R},\gamma)$ 
where $\mathcal{S}$ is the state space, $\mathcal{A}$ is the action space, $\mathcal{P}: \mathcal{S}\times \mathcal{A}\rightarrow \mathcal{S}$ is the MDP transition probabilities, $\mathcal{R}:\mathcal{S}\times\mathcal{A}\rightarrow \mathbb{R}$ is the reward function, and $\gamma\in(0,1)$ is the discount factor.
With each interaction $i$ with the environment $\mathcal{E}$, the agent observes the data tuple $\mathcal{D}_i = (s_i, a_i, s_{i+1}, r_i)$ where $s_i, s_{i+1} \in \mathcal{S}$ is the current and next state, $a_i \in \mathcal{A}$ is the action taken at state $s_i$, and $r_i = \mathcal{R}(s_i, a_i)$ is the reward obtained.  Given a policy $\pi$, the value function (for a given state) and the Q-function (for a given state-action pair) is given by
\begin{align}
    V^{\pi}\left(s \right)=\mathbb{E}\left[\sum_{k=0}^{\infty} \gamma^{k} \mathcal{R}\left(s_{k}, a_{k}\right) \mid s_{0}=s \right] \quad, a_{k} \sim \pi(\cdot|s_{k})
\end{align}
\begin{align}
    Q^{\pi}\left(s, a\right)=\mathbb{E}\left[\sum_{k=0}^{\infty} \gamma^{k} \mathcal{R}\left(s_{k}, a_{k}\right) \mid s_{0}=s, a_{0}=a\right]
\end{align}

We denote by $\rho$ the initial state distribution of state-space $\mathcal{S}$. The long-term discounted return under $\rho$ can be written as $V^{\pi}\left(\rho\right)=\mathbb{E}_{s \sim \rho} V^{\pi}\left(s\right) .$
The goal of RL is to learn an optimal policy $\pi^*$ given the observed data tuples $\mathcal{D}_i$ that maximizes this return i.e. 
\begin{equation}
    \pi^* = \mathop{\arg\!\max}_{\pi} V^{\pi}(\rho)
\end{equation}


We use Q-learning based RL where bellman backup operator is used to update the Q-function until it converges.
\begin{align}
    Q\left(s_i, a_i\right)\leftarrow \mathbb{E}_{s^{\prime} \sim \mathcal{P}(\cdot \mid s_i, a_i)}\left[\mathcal{R}(s_i, a_i)+\gamma \max _{a^{\prime}} Q\left(s_{i+1}, a^{\prime}\right)\right]
    \label{eq:bellman}
\end{align}
\\
and the underlying policy at any iteration is given by
\begin{align}
    a_k =\mathop{\arg\!\max}_{a^{\prime}} Q(s_k, a^{\prime})
\end{align}

The above policy converges to an optimal policy $\pi^*$ (deterministic) when the Q-function converges under the application of the Bellman backup operator.
In order to estimate the Q-function, we consider both the tabular method (for simpler problems) and neural network-based function approximation (for complex problems). In tabular methods, we maintain a Q table of size $|\mathcal{S}|\times |\mathcal{A}|$. 
For more complex problems with a larger state/action space, we use neural network-based function approximation $f_{\theta}: \mathcal{S} \rightarrow \mathcal{A}$ to estimate the Q-function parameterized by $\theta$, where $\theta$ are the trainable weights of the pre-defined neural network structure. 
\subsection{Fault Model}
\begin{figure}

\subfigure[Low obstacle density]{
\hfill\begin{minipage}[b]{0.3\linewidth}
\centering
\includegraphics[width=.8\textwidth]{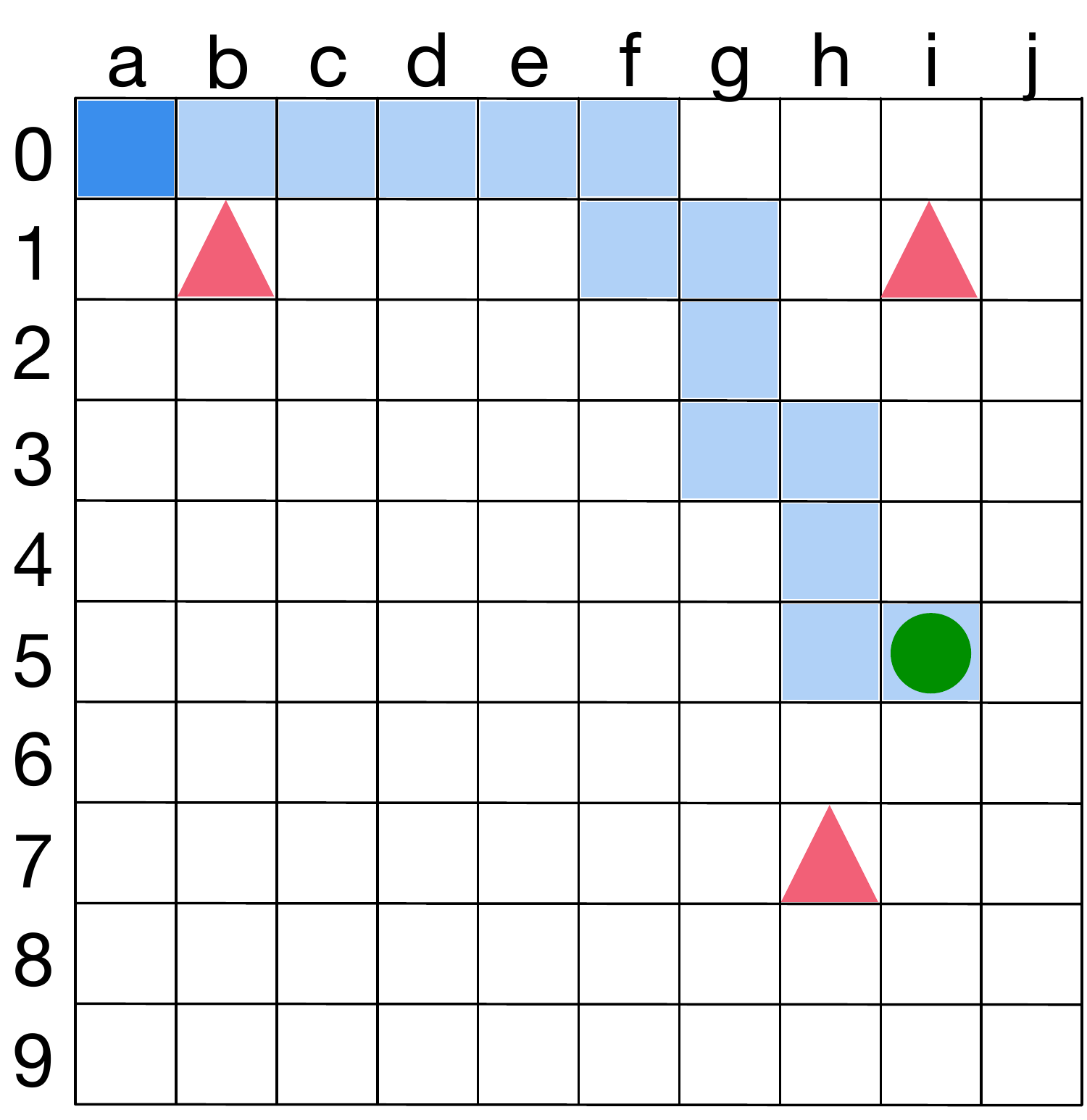}
\end{minipage}
}
\subfigure[Middle obstacle density]{
\begin{minipage}[b]{0.3\linewidth}
\centering
\includegraphics[width=.8\textwidth]{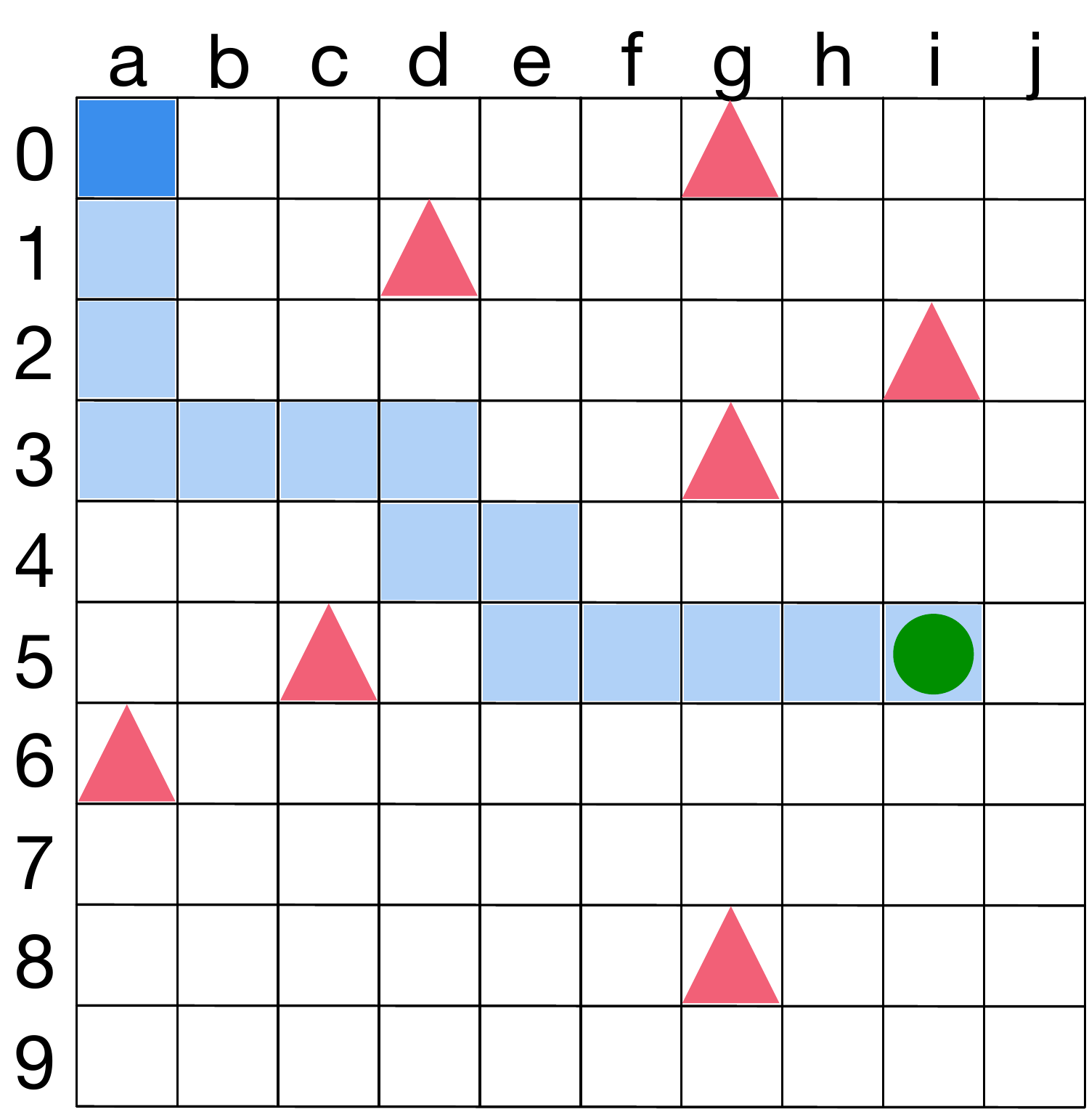}
\end{minipage}
}
\subfigure[High obstacle density]{
\begin{minipage}[b]{0.3\linewidth}
\centering
\includegraphics[width=.8\textwidth]{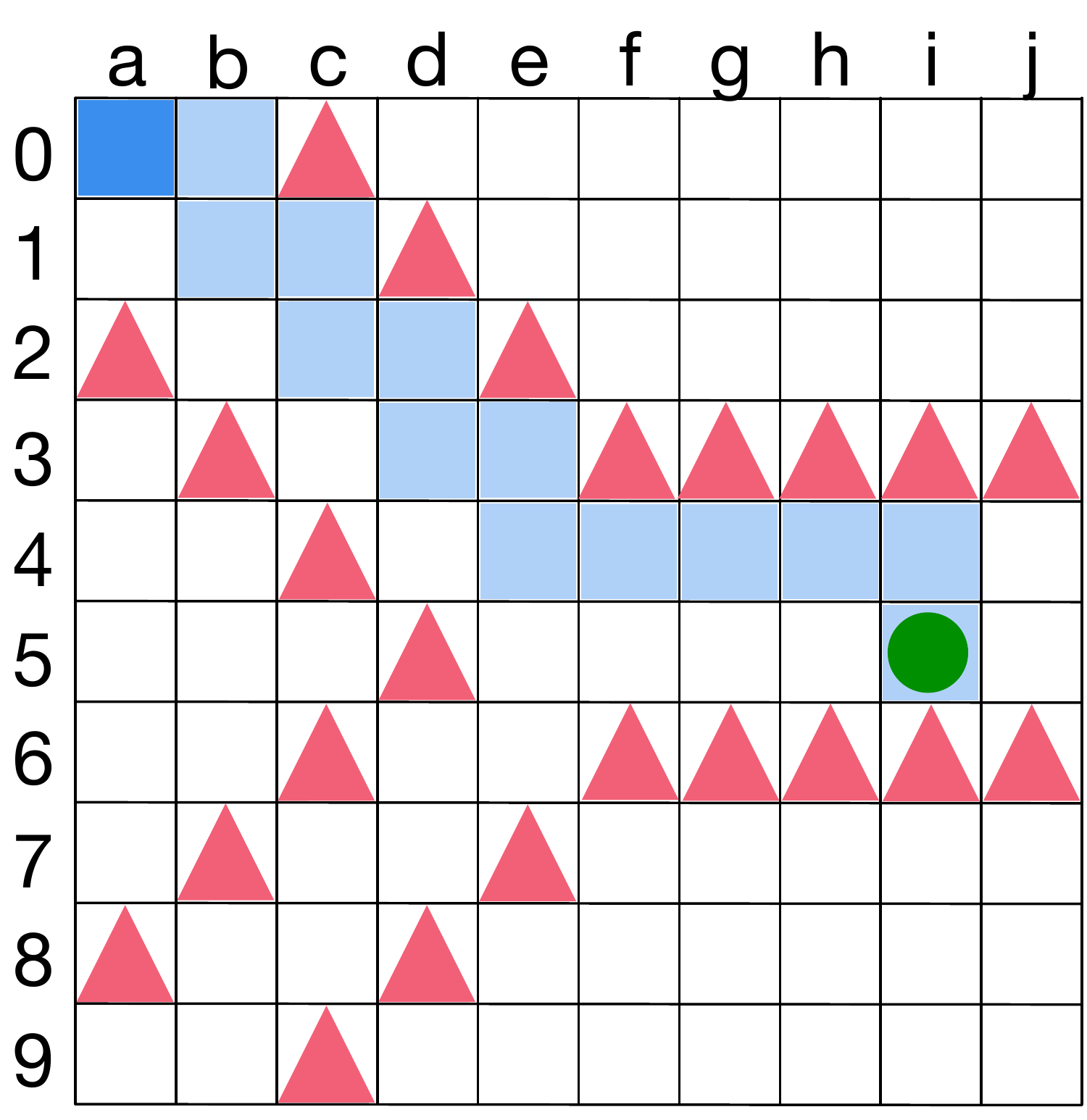}
\end{minipage}
}
\caption{Grid World problems with various obstacle densities. Red, green and dark blue cells are obstacles, goal and agent respectively. Light blue path is the mostly followed route by agent after training.} \label{fig:gd_layout}
\vspace{0.1in}
\end{figure}

\noindent \textbf{Fault type.} We consider both permanent and transient faults in this paper. Permanent faults usually result from manufacturing defects at the transistor and interconnected structures in the integrated circuit and are continuous and stable with time. They can cause bits to be held exclusively low (i.e., stuck-at-0) or high (i.e., stuck-at-1), and introduce bit-flips if the logical value of faulty bits differ from the value where the stuck happens. Transient faults (i.e., soft error) mainly arise from external perturbations, such as particle strikes or voltage violations, and may only exist for a short period. They manifest as bit-flips in the computing and storage elements and result in incorrect random values. We use widely adopted stuck-at and random bit-flip fault models as abstractions of permanent and transient physical defect mechanisms~\cite{salami2018resilience}.

\noindent \textbf{Fault location.}
We consider faults in the memory, which are pertinent to the learning-based system due to its large intermediate states and values storage requirement. For tabular-based policy, we consider data buffer storing tabular values. For neural network-based policy, we consider input buffer, filter buffer and output buffer which are used to store feature maps, weights and activations, respectively. Faults can also happen in the datapath, and we assume that faults in MAC units end up corrupting the operation outputs and are manifested as random values in the output buffer. We do not consider faults in control logic because the scheduling is mainly done by the host (i.e., CPU). The fault model is derived from~\cite{reagen2018ares,li2017understanding} and interested readers are pointed to the references for details.

\subsection{Fault Injection Method}
Fault injection is performed in two modes: static and dynamic. Static faults are injected offline before training and before inference is executed. Dynamic injection is conducted during the execution, and the overheads are minimized by implementing the fault model as tensor operations. This method is in line with~\cite{reagen2018ares} and its accuracy has been validated on silicon~\cite{whatmough201714}. Permanent faults are injected statically since they are independent of execution. Transient faults in weights during inference are static injection since they are known after training and can be manipulated before inference begins. In contrast, transient faults in activations are injected dynamically since they are input-dependent.

\section{Fault Characterization}
\label{sec:experiments}

In this section, we evaluate the impact of faults on two learning-based template problems: Grid World (Sec.~\ref{subsec:gd}) and drone autonomous navigation (Sec.~\ref{subsec:PEDRA}). 

\subsection{Grid World Navigation Problem}
\label{subsec:gd}
We begin our experimentation with a simpler but popular navigation problem of Grid World. The environment consists of a $n\times n$ grid. Each cell in the grid is marked as either of the four categories: \{\texttt{source}, \texttt{goal}, \texttt{hell}, \texttt{free}\}. The agent is placed at the \texttt{source} and is required to reach the \texttt{goal} and avoid getting trapped in  \texttt{hell}. Each cell corresponds to a unique state ($|\mathcal{S}|=n^2$). At each occupied cell, the agent samples from one of the four actions from its action space $\mathcal{A}$: \{\texttt{move-up}, \texttt{move-down}, \texttt{move-left}, \texttt{move-right}\} ($|\mathcal{A}|=4$) and moves to a new cell. The reward is 1, -1 or 0 if the new cell is a \texttt{goal}, \texttt{hell} or \texttt{free} respectively. The performance of policy is quantified by the agent's \texttt{success-rate} which is defined as the ratio of the number of times the agent reached the \texttt{goal} to the total number of trials and cumulative reward (sum of rewards in an episode).
We consider three Grid World settings with various obstacle densities, as shown in Fig.~\ref{fig:gd_layout}. Widely used tabular-based and neural network (NN)-based methods (Sec.~\ref{subsec:RL_methodology}) are adopted and compared. The policies are quantized to 8-bit without loss of performance during the training and inference given the memory and power constraints of edge devices. We repeat each fault injection campaign 1000 times, which can lead to 95\% confidence-level within 1\% error margin. Due to space limit, we only show results for Fig.~\ref{fig:gd_layout}b. We however confirm that similar observations apply to both Fig.~\ref{fig:gd_layout}a and \ref{fig:gd_layout}c.

\subsubsection{Training in Grid World}
\begin{figure}
\centering
\subfigure[Tabular-based approach]{
\begin{minipage}[b]{0.57\linewidth}
\includegraphics[width=\textwidth]{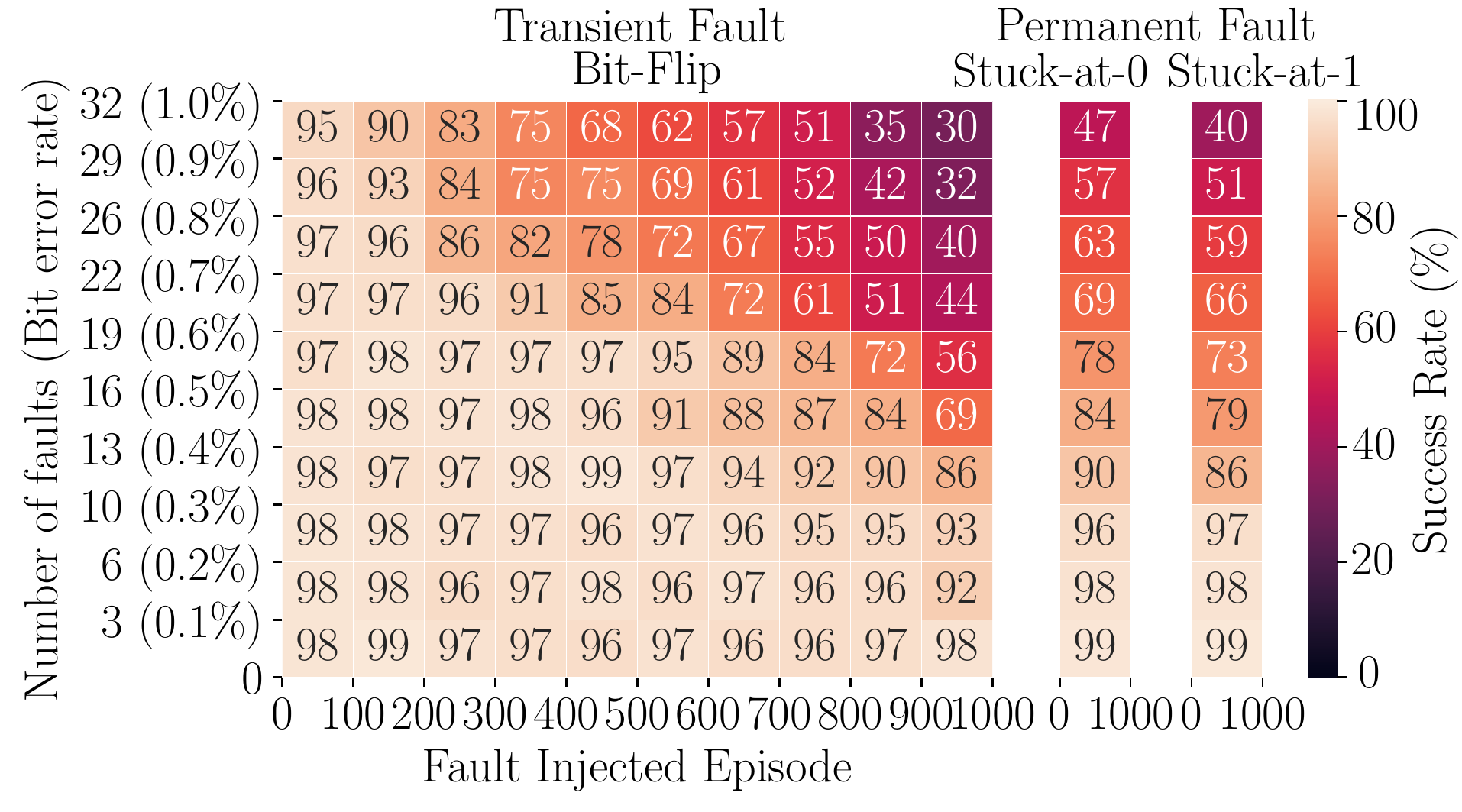}
\end{minipage}
}
\hspace{-0.14in}
\subfigure[Tabular value histogram]{
\begin{minipage}[b]{0.41\linewidth}
\includegraphics[width=\textwidth]{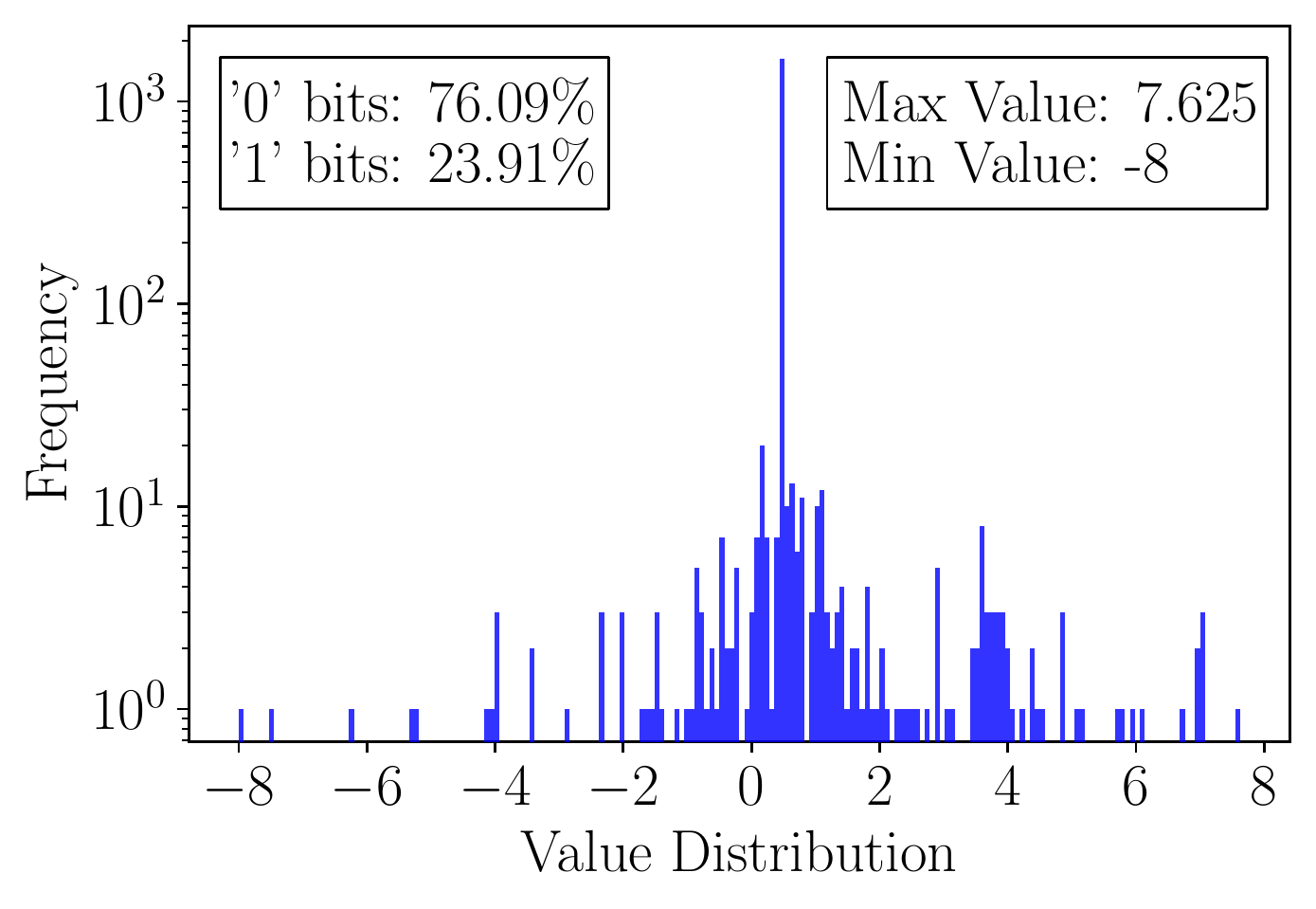}
\end{minipage}
}\\
\hspace{-0.15in}
\subfigure[NN-based approach]{
\begin{minipage}[b]{0.57\linewidth}
\includegraphics[width=\textwidth]{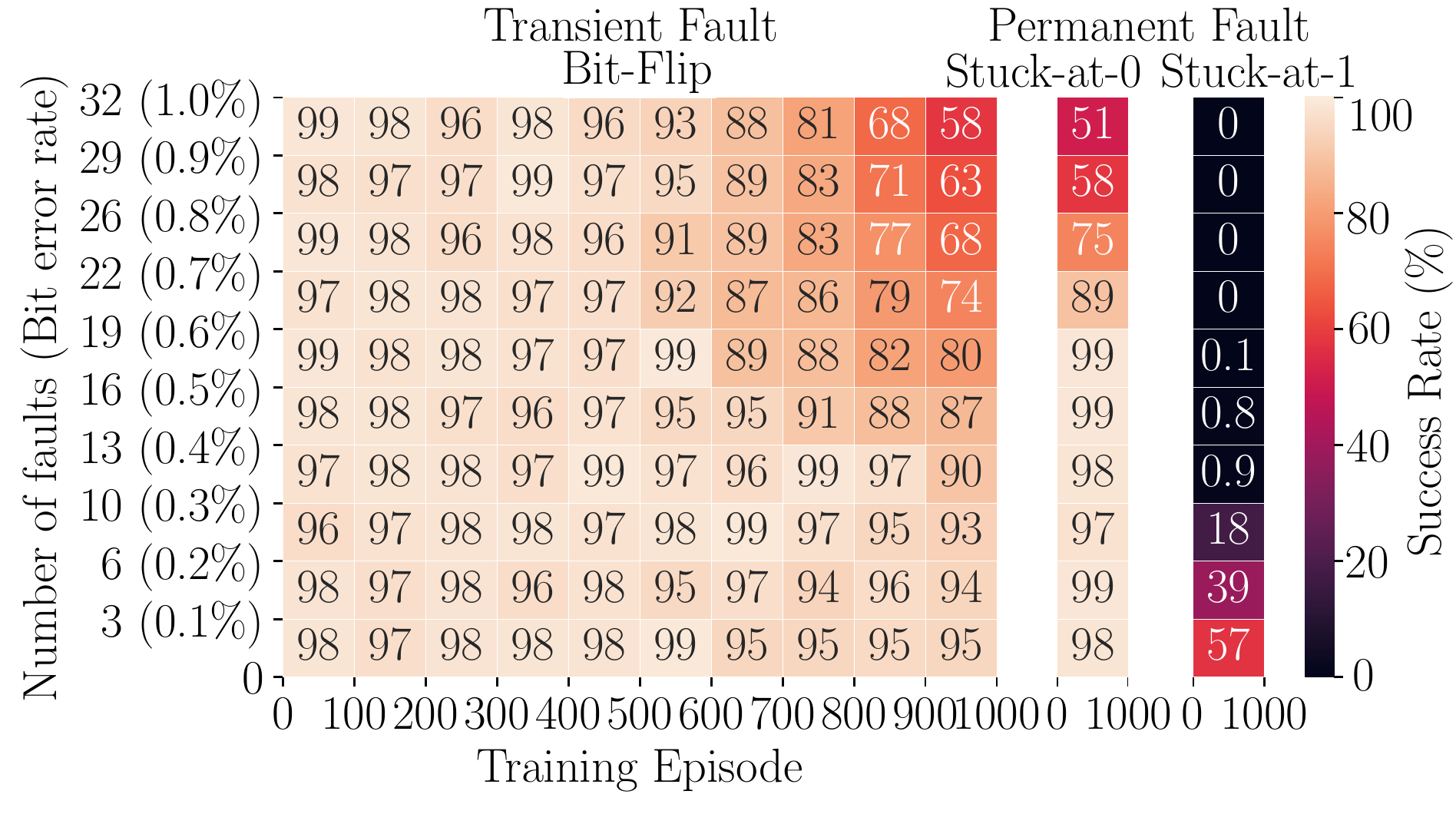}
\end{minipage}
}
\hspace{-0.14in}
\subfigure[NN weight histogram]{
\begin{minipage}[b]{0.41\linewidth}
\includegraphics[width=\textwidth]{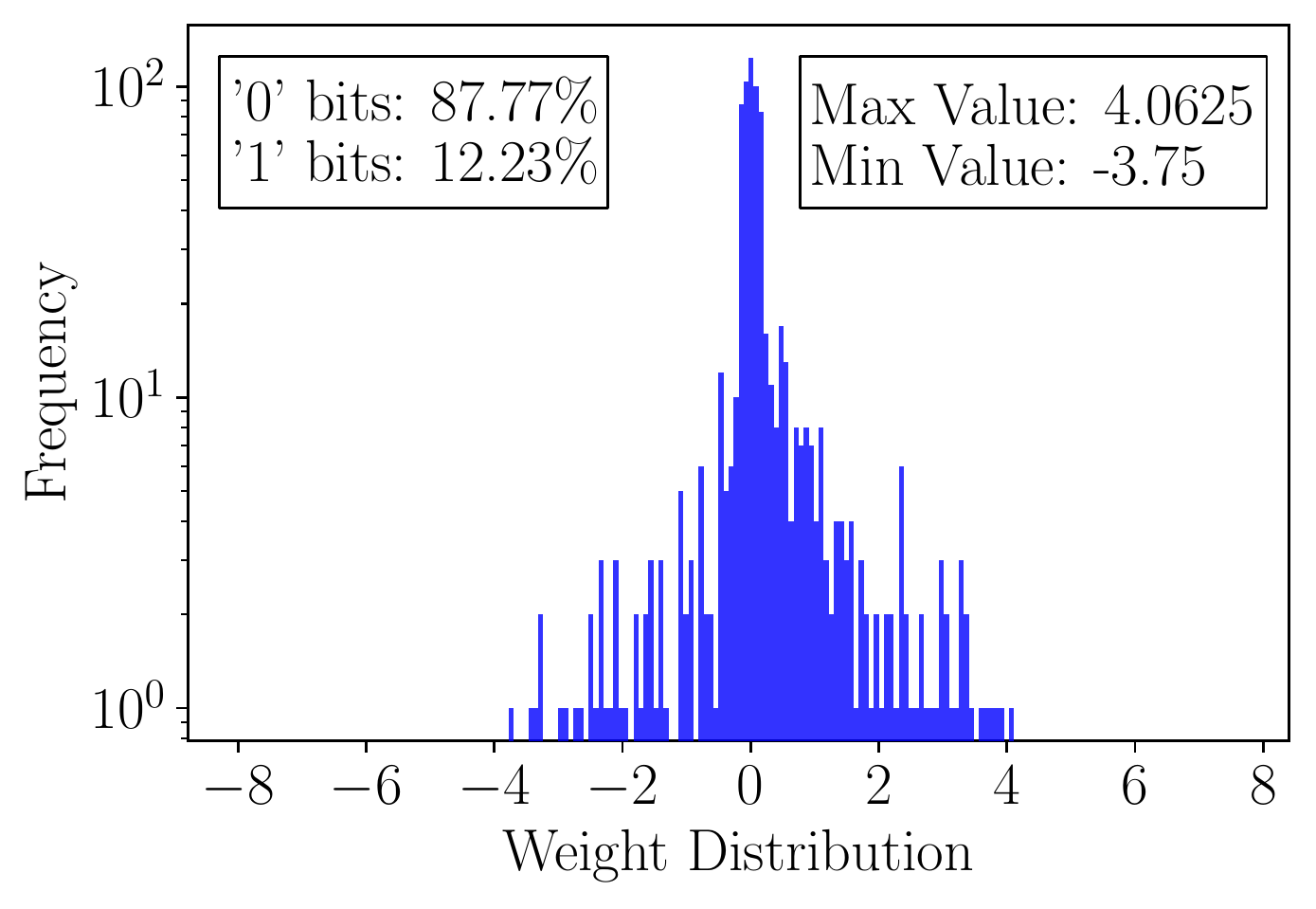}
\end{minipage}
}
\hspace{-0.15in}
\caption{(a)(c) The impact of transient and permanent faults on Grid World training for both tabular and NN-based approaches. The darker, the worse of performance. (b)(d) The histogram of tabular values and NN parameters.} \label{fig:gd_training_heatmap}
\vspace{0.1in}
\end{figure}

\noindent\textbf{\\Transient faults in training.}
\label{subsubsec:transient_gw}
Fig.~\ref{fig:gd_training_heatmap}a and Fig.~\ref{fig:gd_training_heatmap}c show the impact of transient faults on both tabular and NN-based methods during training. Bit-flips are injected in a single random episode with different bit error rates (BERs). It is well observed that transient faults occurred in earlier episodes with low BERs have a negligible impact on the agent's final success rate (color-coded). However, when the BER and fault injection episode are introduced beyond a certain threshold, the performance and accuracy degrades quickly.


We find that with low BER, the agent can learn from these faults during training and generate a good policy. In a few cases, policy trained with bit-flips even yields a higher success rate than runs without faults. We believe that these bit-flips can be seen as injecting an amount of noise that was small enough to maintain a good policy and large enough to improve exploration which benefits the convergence. This is akin to regularization in RL and previous work in quantization report similar effect~\cite{krishnan2019quantized}. However, with high BER, the agent fails to learn from these faults which decreases the efficacy of the policy and can even impact the convergence.

\begin{figure}
\centering
\subfigure[Tabular-based approach]{
\begin{minipage}[b]{0.48\linewidth}
\includegraphics[width=\textwidth]{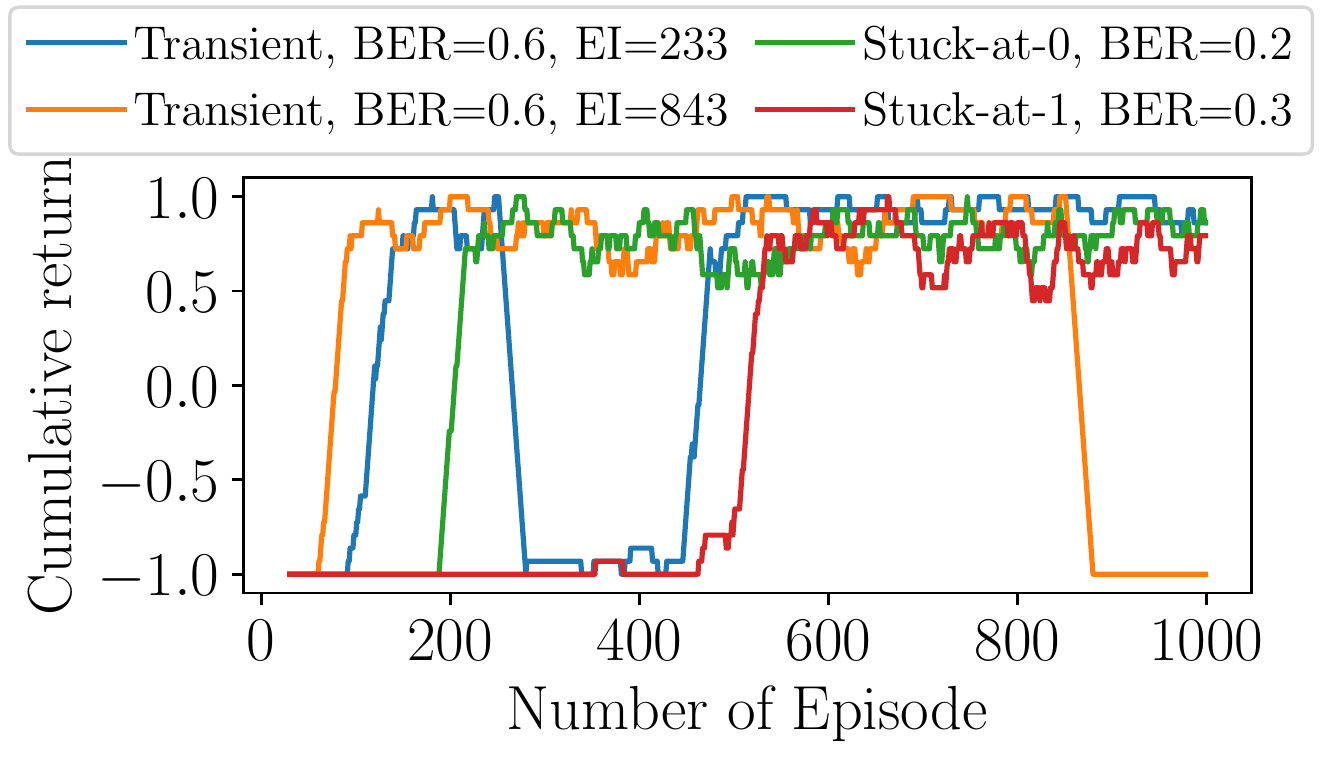}
\end{minipage}
}
\hspace{-0.08in}
\subfigure[NN-based approach]{
\begin{minipage}[b]{0.48\linewidth}
\includegraphics[width=\textwidth]{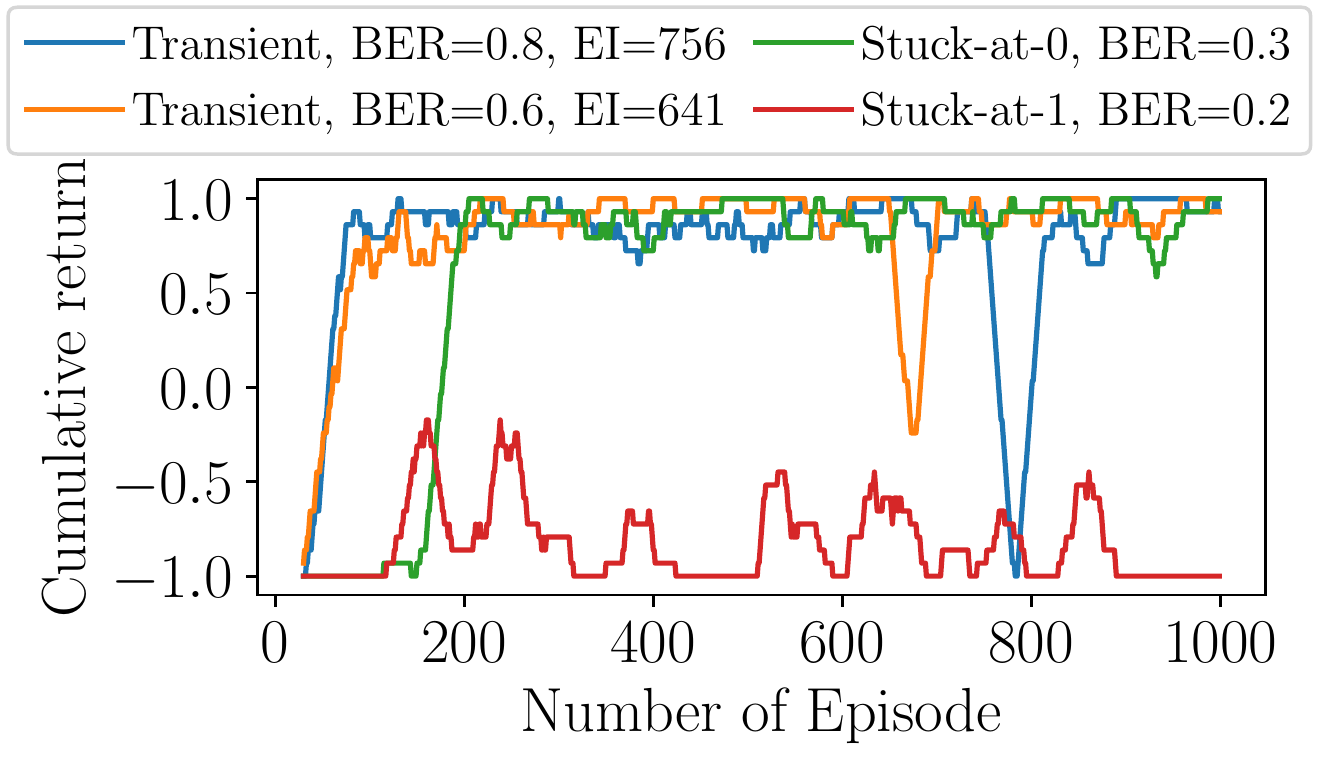}
\end{minipage}
}
\caption{Exampled cumulative return during training under transient and permanent faults. BER represents bit error rate (\%). EI represents faults injected episode index.} \label{fig:training_reward}
\end{figure}

Another takeaway from Fig.~\ref{fig:gd_training_heatmap}a and Fig.~\ref{fig:gd_training_heatmap}c is that NN-based approach exhibits higher resilience to transient faults than the tabular approach. We hypothesize the following two reasons: NN-based training process is not significantly impacted by transient faults and it can quickly recover from faults. Fig.~\ref{fig:training_reward} shows that the latter case contributes more to the policy resiliency. After bit-flips get injected (blue and orange curves), the cumulative rewards of both tabular and NN-based training drop quickly, indicating that faults destroy existing learned experience. However, the NN-based approach recovers from faults quickly after training several episodes due to its advanced optimization techniques, whereas the tabular approach takes a longer time to recover or even fails to converge at the end of the training.

\noindent \textbf{Permanent faults in training.}
Permanent fault is stuck at 0 or 1 at the same memory location throughout training, and its impacts are shown in Fig.~\ref{fig:gd_training_heatmap}a and Fig.~\ref{fig:gd_training_heatmap}c. We find that despite its permanent nature, these fault have comparable impact as transient faults on the policies. This indicates that the agent can learn and adapt itself to the permanent fault patterns in training. For example, if bits are stuck at 0 and affect a cell which will be passed with the uncontaminated policy, the agent will gradually learn to avoid this cell and find a new path. Due to system's intrinsic long-term learning capability, permanent faults do not corrupt the learned policies as much as transient faults that occur in the latter training episodes.

Interestingly, stuck-at-1 has a slightly higher impact than stuck-at-0 on policy performance for tabular-based approach, as shown in Fig.~\ref{fig:gd_training_heatmap}a. This can be explained from Fig.~\ref{fig:gd_training_heatmap}b where tabular values contain more 0 bits than 1 bits, therefore stuck-at-1 bring more faulty bits in the system. But due to the tabular values' characteristics (not centered at 0 and a wide range) and cell's functionality (most cells don't contribute to the final path and one cell only affects its adjacents), the distinction between the impact of stuck-at-0 and stuck-at-1 faults is small.

Different from tabular method, stuck-at-0 and stuck-at-1 exhibit significantly divergent impact on NN-based methods. As shown in Fig.~\ref{fig:gd_training_heatmap}c, stuck-at-0 has a negligible impact within 0.7\% BER, while stuck-at-1 noticeably fails the training even with 0.2\% BER. This observation is due to the sparsity and limited range of NN data as visualized in Fig.~\ref{fig:gd_training_heatmap}d. At a bit level, there are more 0s than 1 bits in the weights by 7.17$\times$, whereas this ratio is only 3.18$\times$ for tabular values. Further, most NN model data is centered around 0 and have a much narrower range than tabular values. From NN characteristics, weights with higher magnitude play more critical roles in final output. Therefore, the greater magnitude faulty values arose from stuck-at-1 have more impacts on the system than stuck-at-0.

The training processes under permanent faults are demonstrated in Fig.~\ref{fig:training_reward} (green and red curves). The agent can learn across stuck-at-faults and finally converge itself although it takes a longer time in the beginning phase of training, except for stuck-at-1 in the NN-based method which destroys the convergence of system.

\begin{figure}
\centering
\subfigure[Transient faults in Tabular-based training]{
\begin{minipage}[b]{0.35\linewidth}
\includegraphics[width=\textwidth]{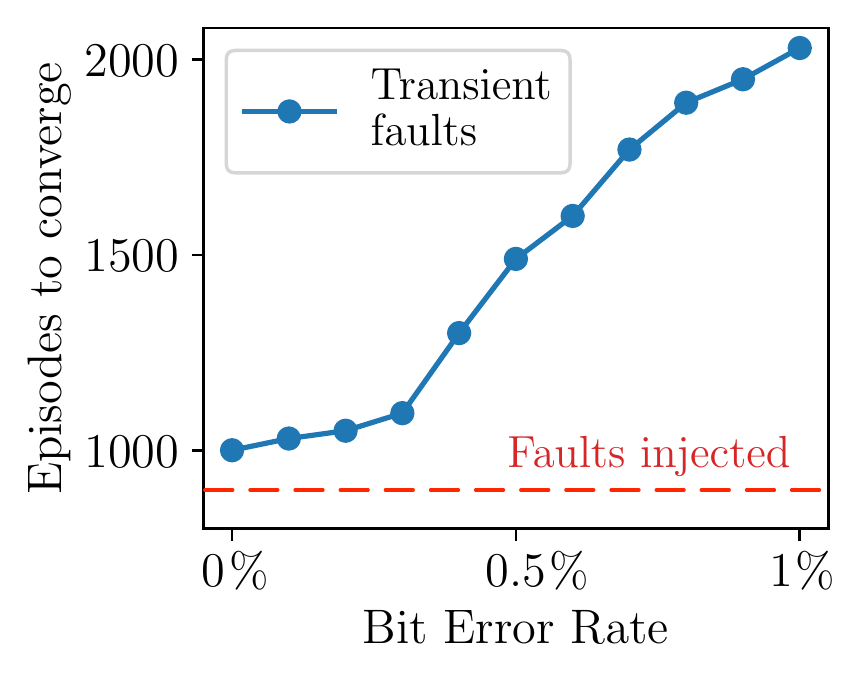}
\end{minipage}
}
\hspace{-0.08in}
\subfigure[Permanent faults in Tabular-based training]{
\begin{minipage}[b]{0.58\linewidth}
\includegraphics[width=\textwidth]{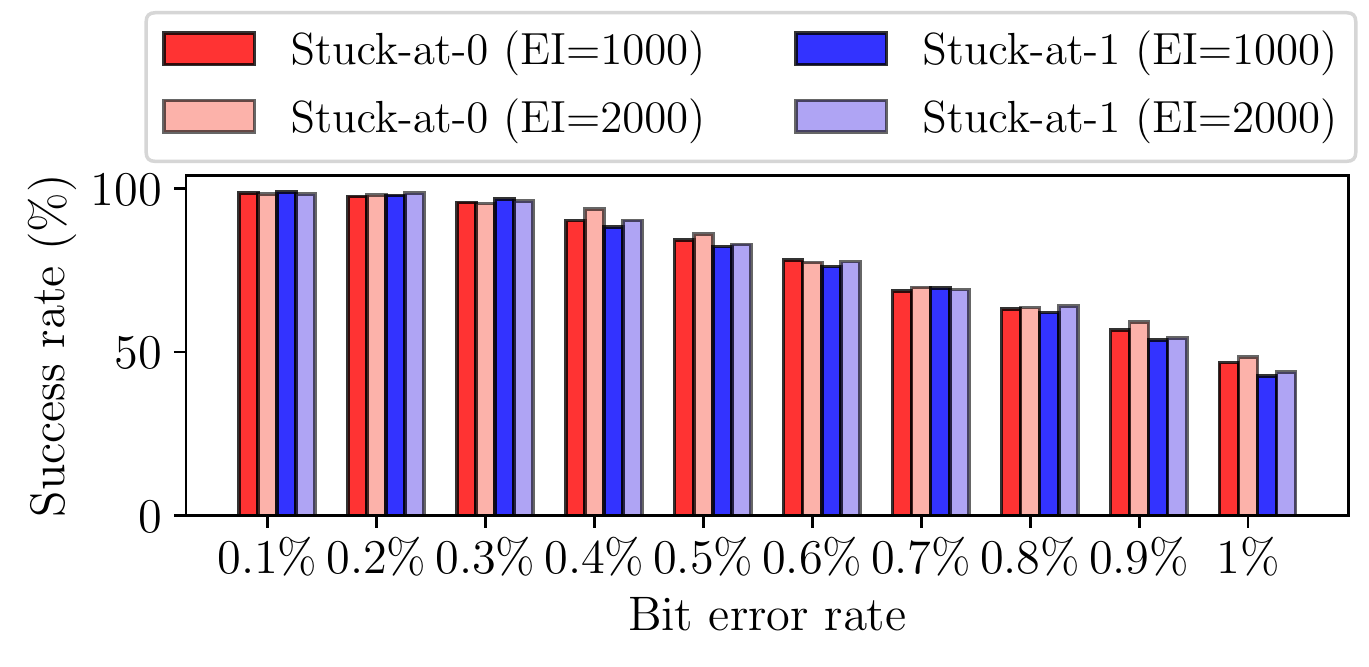}
\end{minipage}
}\\
\subfigure[Transient faults in NN-based training]{
\begin{minipage}[b]{0.35\linewidth}
\includegraphics[width=\textwidth]{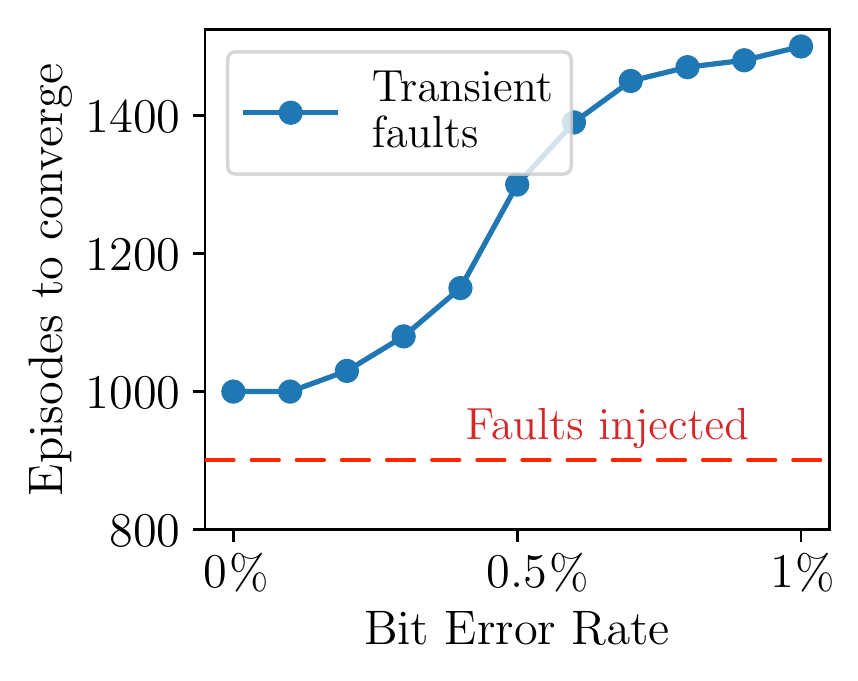}
\end{minipage}
}
\hspace{-0.08in}
\subfigure[Permanent faults in NN-based training]{
\begin{minipage}[b]{0.58\linewidth}
\includegraphics[width=\textwidth]{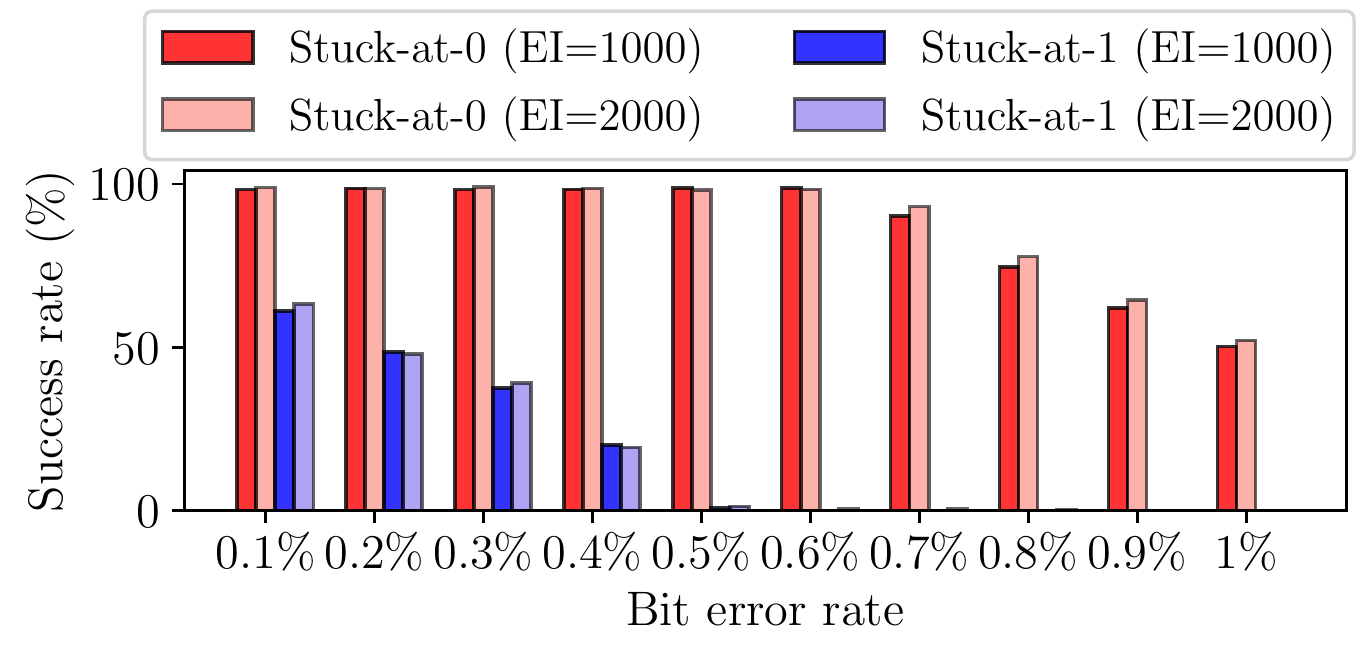}
\end{minipage}
}
\caption{(a)(c) Episodes taken to converge ($>$95\% success rate) after transient faults injected in 900th episode. (b)(d) The policy's success rate after training 1000 more episodes under permanent faults. EI represents faults injected episode index.} \label{fig:gd_converge}
\vspace{0.1in}
\end{figure}

\noindent \textbf{Convergence.}
We now ask the following question: whether the policy can finally converge after faults are injected? Answering this question will enable us to further understand the system's long-term learning capability on different types of faults.
Fig.~\ref{fig:gd_converge}a and Fig.~\ref{fig:gd_converge}c demonstrate that both tabular and NN-based approach can finally achieve convergence after transient faults injected, although tabular method needs 2$\times$ episodes than NN method. This further implicates the self-healing ability and higher resiliency to transient faults of NN-based learning system.
In contrast, Fig.~\ref{fig:gd_converge}b and Fig.~\ref{fig:gd_converge}d indicate that extra training doesn't bring obvious improvements for both stuck-at-0 and stuck-at-1 when BER surpasses a threshold, and may even degrade the performance due to over-training. This is because a steady low exploration rate impedes the agent learn the permanent fault, which we will discuss more in Sec.~\ref{sec:faullt_mitigatioin}.

\subsubsection{Inference in Grid World}
\begin{figure}
\centering
\subfigure[Tabular-based inference]{
\begin{minipage}[b]{0.47\linewidth}
\includegraphics[width=\textwidth]{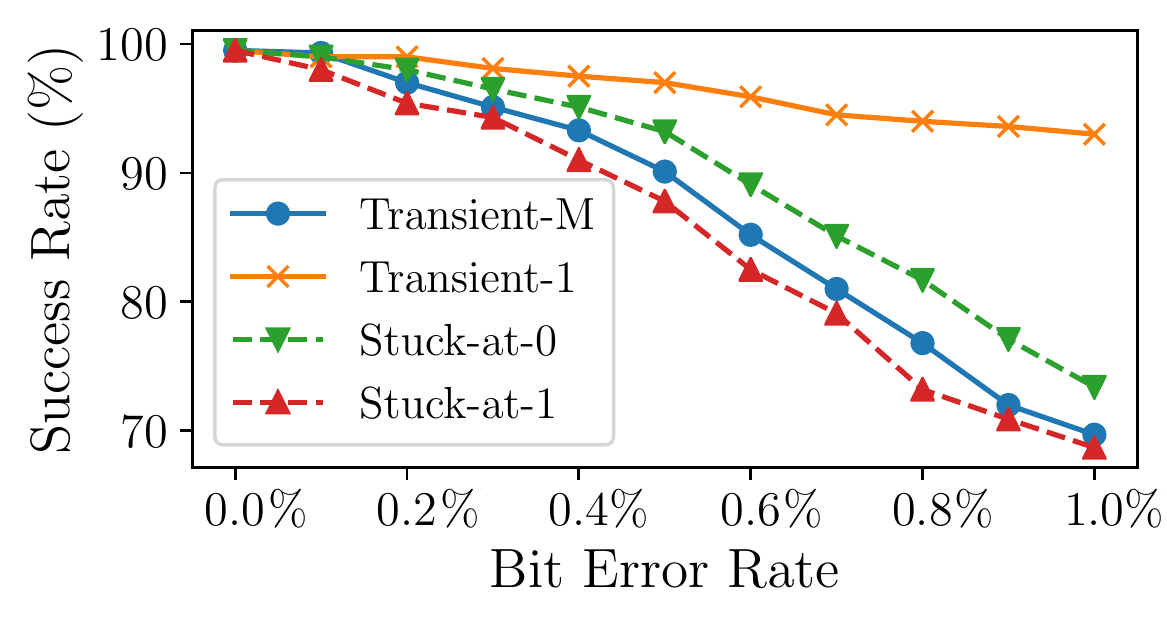}
\end{minipage}
}
\hspace{-0.06in}
\subfigure[NN-based inference]{
\begin{minipage}[b]{0.47\linewidth}
\includegraphics[width=\textwidth]{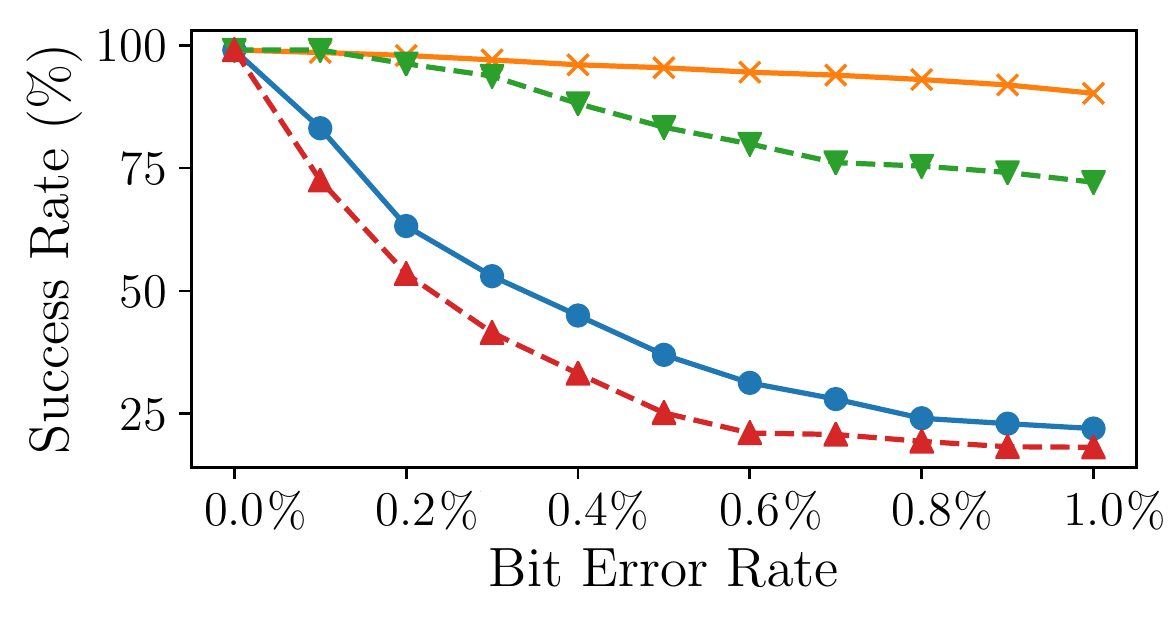}
\end{minipage}
}
\caption{The impact of transient and permanents faults on Grid World inference for both tabular and NN-based approaches.} \label{fig:gd_inference}
\end{figure}

\noindent \textbf{\\Transient faults in inference.} Inference of learning-based system is a sequential decision-making process. The trained policy will be utilized in each action step based on the agent's current state. Hence transient faults have two modes: occur in the read register which only impacts the current single step (Transient-1) or occur in memory which impacts the whole executions (Transient-M). As shown in Fig.~\ref{fig:gd_inference}, Transient-1 has a negligible effect on navigation success rate compared to transient-M fault, this is because a single wrong decision doesn't necessarily results in task failure and can be remedied in the following actions of learning-based system.

\noindent \textbf{Permanent faults in inference.} Permanent faults are stuck at 0 or 1 during whole inference steps. For tabular-based inference, stuck-at-0 and stuck-at-1 have similar impacts. For NN-based inference, however, stuck-at-1 has a much severe impact on policy than stuck-at-0 due to the sparsity nature of NN parameters.



\begin{figure}
\centering
\subfigure[PEDRA environments]{
\begin{minipage}[b]{0.48\linewidth}
\includegraphics[width=\textwidth]{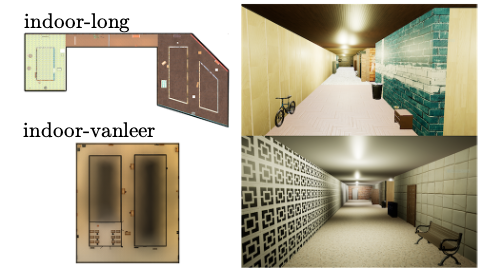}
\end{minipage}
}
\hspace{-0.05in}
\subfigure[C3F2 neural network policy]{
\begin{minipage}[b]{0.48\linewidth}
\includegraphics[width=\textwidth]{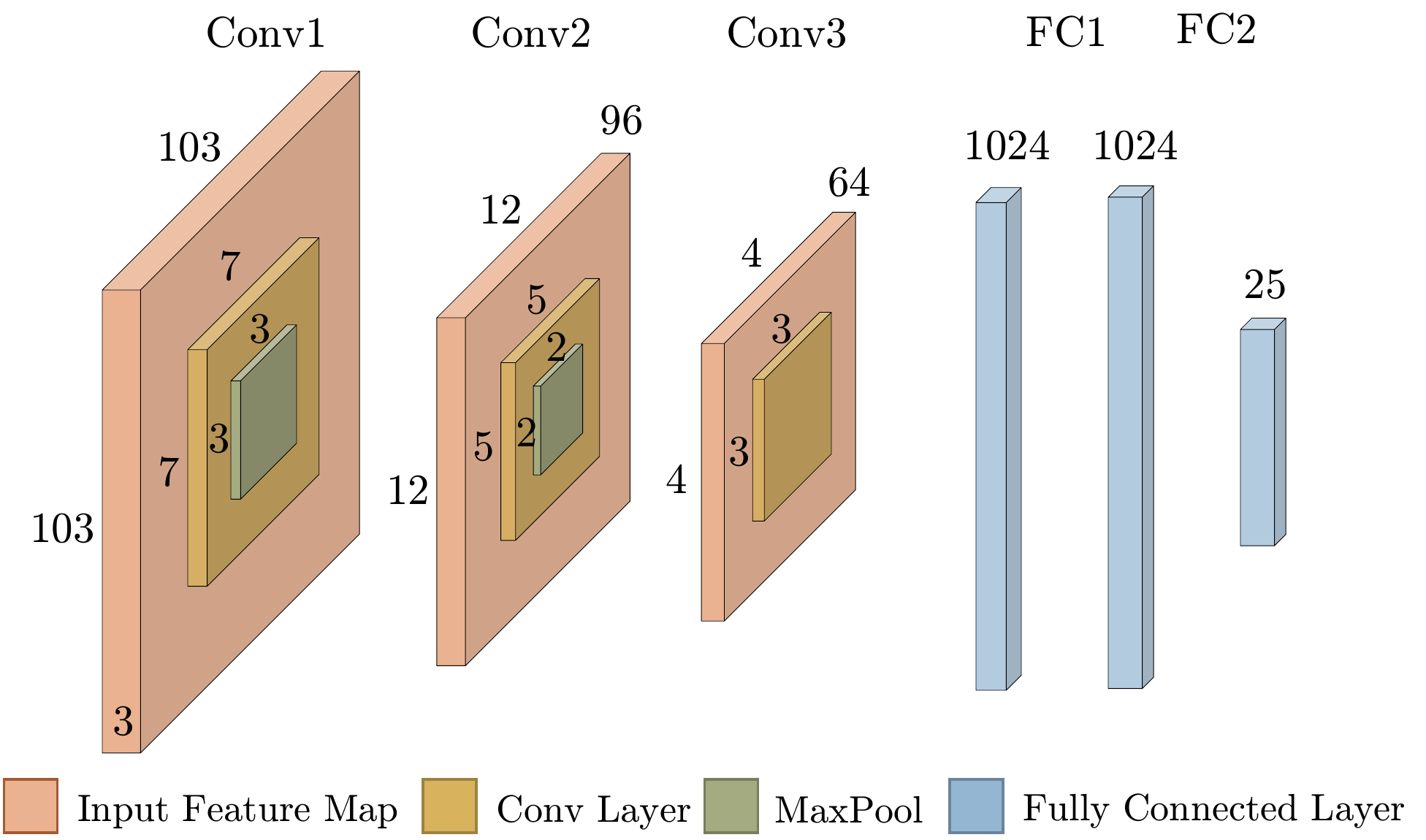}
\end{minipage}
}
\caption{Training details for the drone autonomous navigation task} \label{fig:pedra}
\vspace{0.1in}
\end{figure}

\begin{figure*}
\centering
\subfigure[Drone navigation training]{
\begin{minipage}[b]{0.34\linewidth}
\includegraphics[width=\textwidth]{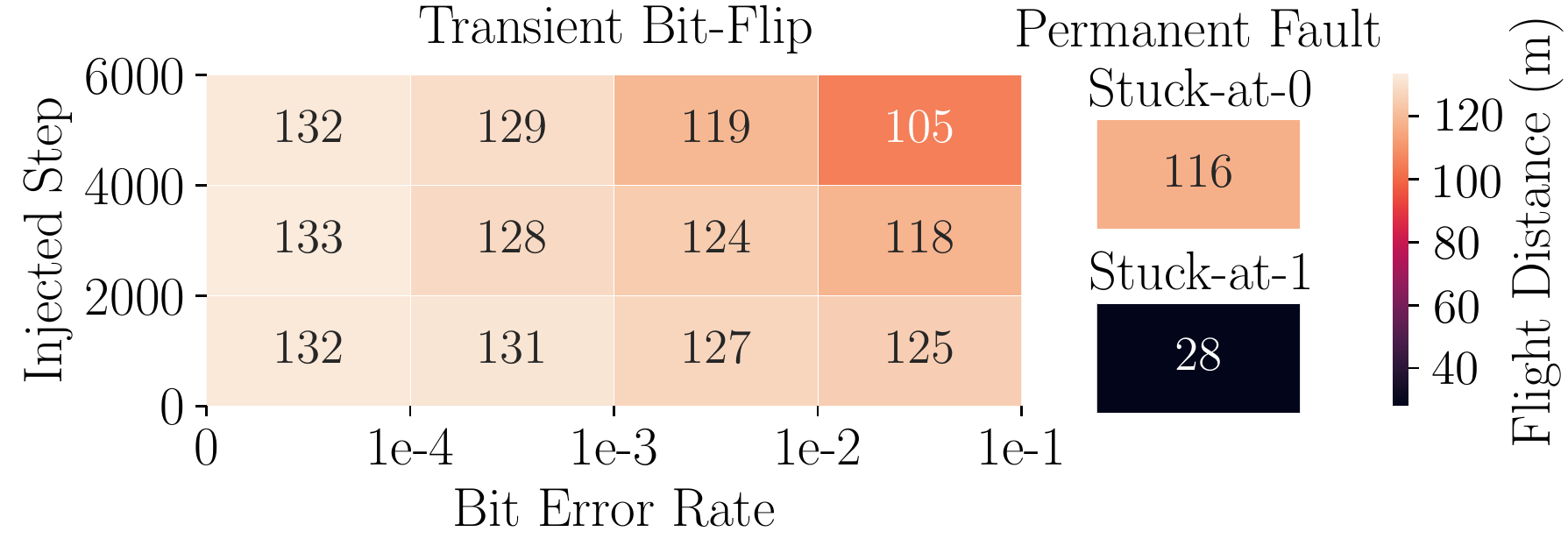}
\end{minipage}
}
\hspace{-0.05in}
\subfigure[Environment]{
\begin{minipage}[b]{0.26\linewidth}
\includegraphics[width=\textwidth]{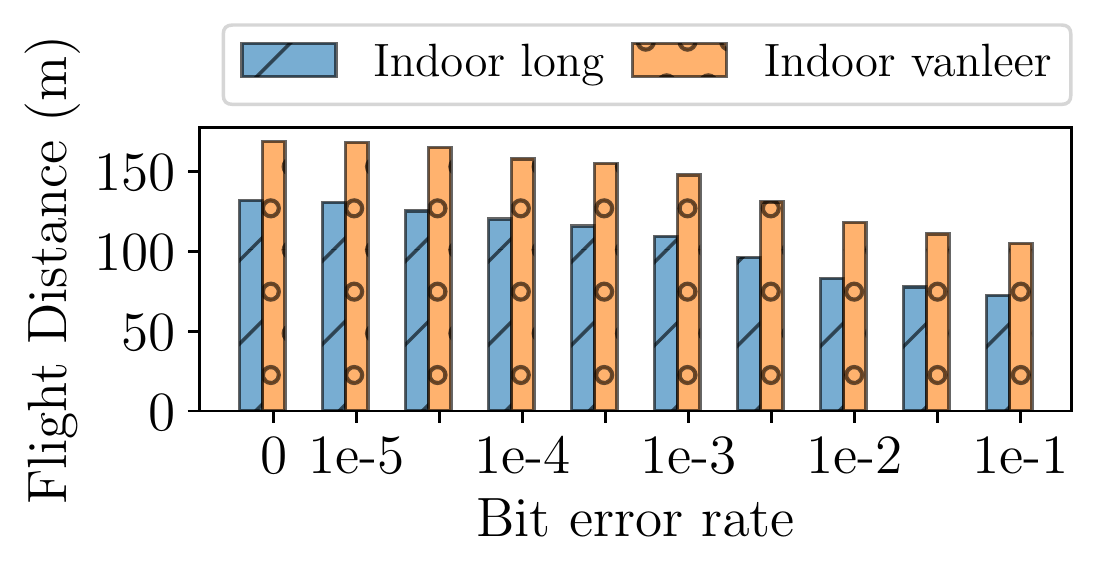}
\end{minipage}
}
\hspace{-0.12in}
\subfigure[Data location]{
\begin{minipage}[b]{0.34\linewidth}
\includegraphics[width=\textwidth]{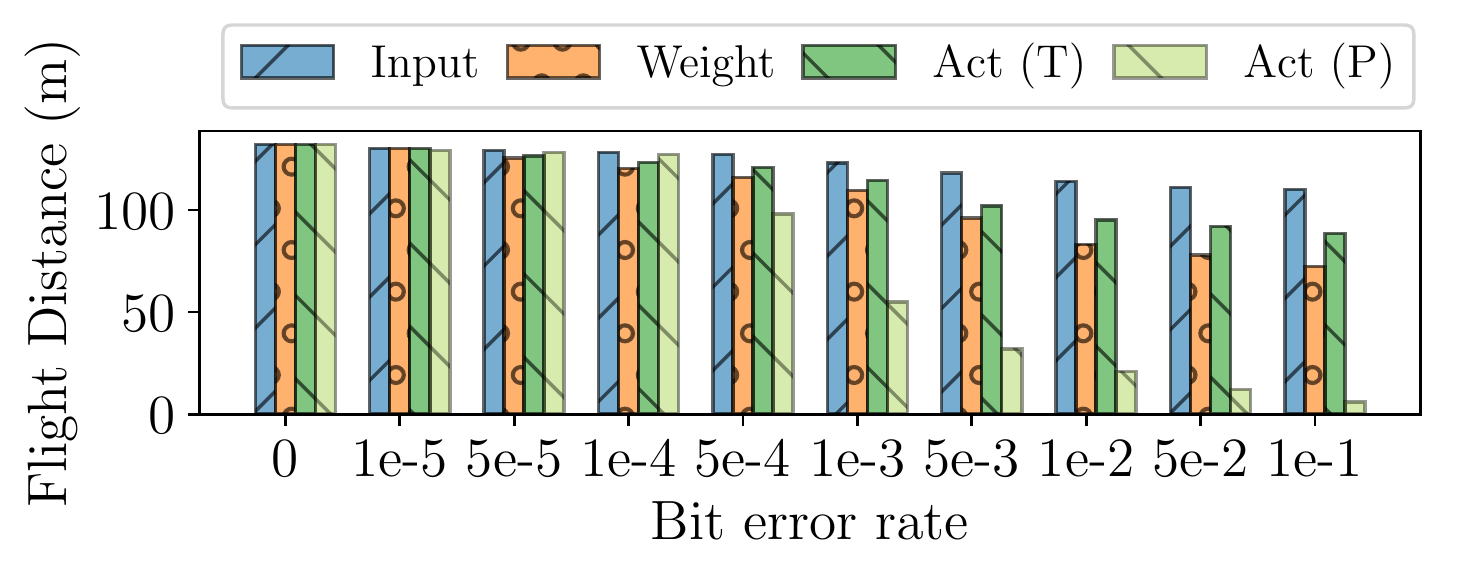}
\end{minipage}
}\\
\subfigure[NN layer]{
\begin{minipage}[b]{0.4\linewidth}
\includegraphics[width=\textwidth]{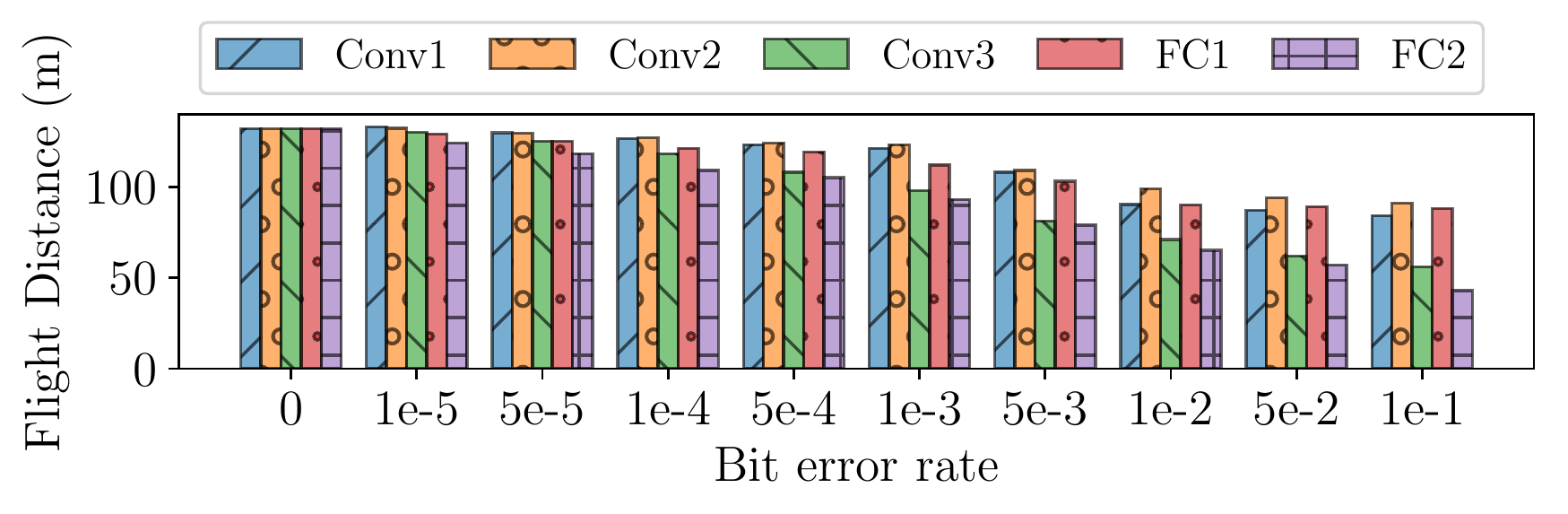}
\end{minipage}
}
\hspace{-0.15in}
\subfigure[Data type]{
\begin{minipage}[b]{0.3\linewidth}
\includegraphics[width=\textwidth]{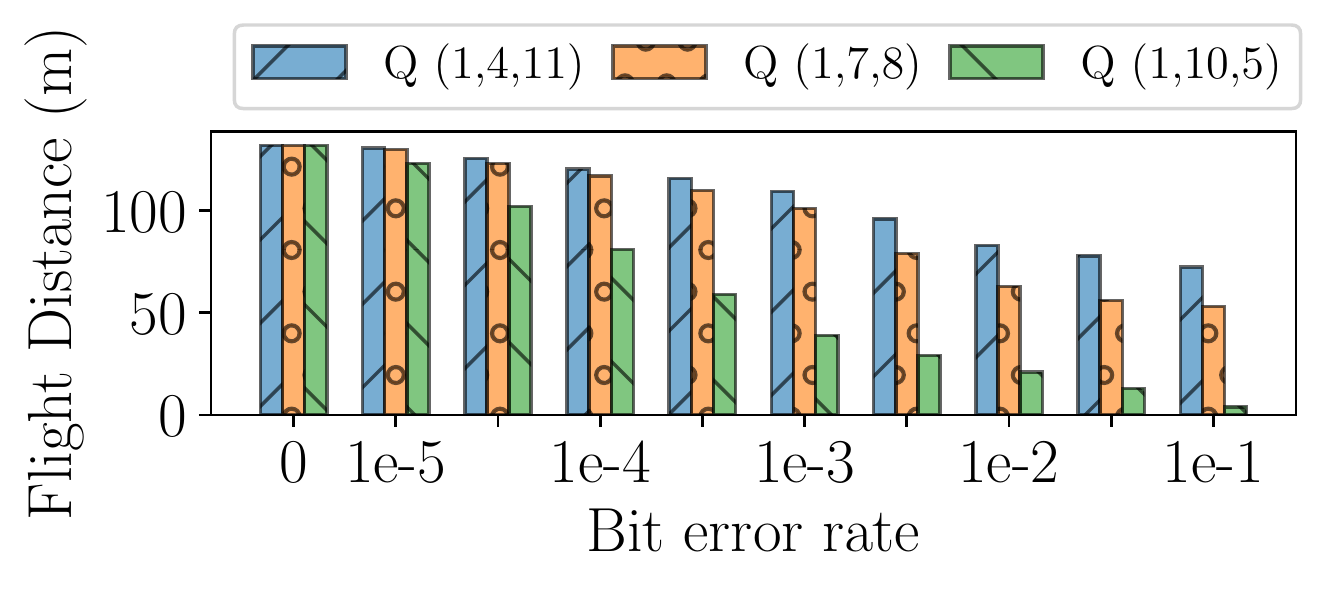}
\end{minipage}
}
\caption{Fault characterization on drone navigation tasks. (a) is training result, (b)-(e) are inference results.} \label{fig:pedra_inference}
\vspace{0.1in}
\end{figure*}

\subsection{Drone Autonomous Navigation}
\label{subsec:PEDRA}

In this section, we experiment on a more complex drone navigation problem in 3D realistic environments. Drone is a complex system~\cite{krishnan2020sky}, and we use PEDRA \cite{anwar2020autonomous} as the simulation platform which is powered by Unreal Engine. During the task, the drone is initialized at a start location and required to navigate across the environment avoiding obstacles. There is no destination position, and the drone is required to fly as long as it can. The monocular image captured from the front-facing camera is taken as the state of the RL problem. We use a perception-based probabilistic action space $\mathcal{A}$ with 25 actions. The reward is designed to encourage the drone to stay away from the obstacles. We use the \texttt{indoor-long} and \texttt{indoor-vanleer} environments provided by PEDRA, which can be seen in Fig. \ref{fig:pedra}a. To quantify the policy performance, we use Mean Safe Flight (MSF) which is the average distance traveled by the drone before collision. 


\subsubsection{Training in Drone Navigation Problem}
\noindent\textbf{\\Transient faults in training.} The model (Fig.~\ref{fig:pedra}b) is first trained offline using Double DQN with experience replay and then fined-tuned last two layers online using transfer learning ~\cite{anwar2020autonomous}. Transient fault is injected in a single random step with various BER during online training, and its impact is shown in Fig.~\ref{fig:pedra_inference}a. We can observe that the flight quality degrades with higher BER and latter fault occurrence, which is the same trend as Sec.~\ref{subsubsec:transient_gw}.


\noindent\textbf{Permanent faults in training.} Permanent faults exist throughout online training. Fig.~\ref{fig:pedra_inference}a demonstrates
that stuck-at-1 faults (BER=$10^{-3}$) severely affect the drone's safe flight distance, while stuck-at-0's impact is moderate. These are the same trends shown in Sec.~\ref{subsubsec:transient_gw}, so we omit the explanation here due to space restriction. 


\subsubsection{Inference in Drone Navigation Problem}
\noindent\textbf{\\Different environments.} We first compare the policy resilience between two environments, shown in Fig.~\ref{fig:pedra_inference}b. Transient faults are injected in weights once the policy is trained. Experiments are repeated 100 times for each case. We observe that faults have similar impact trends on both \texttt{indoor-long} and \texttt{indoor-vanleer}, which brings more flight quality degradation with higher BER. Due to this little difference, we focus on \texttt{indoor-long} for the rest of Fig.~\ref{fig:pedra_inference}.

\noindent \textbf{Different fault locations.} The fault sensitivity of different data locations (i.e., feature map, weights and activations) is evaluated in Fig.~\ref{fig:pedra_inference}c. Weight buffer is read-only once the policy is trained, while input and activation buffer will be rewritten during execution. Fig.~\ref{fig:pedra_inference}c shows that input buffer is relatively immune to transient faults since bit-flips only affect values in a single frame and image information has redundancy. Transient faults occurred in ReLU activation (Act (T)) bring more flight distance drop than inputs because an activation faulty value will be propagated to all neurons in the next level without any masking. Permanent faults occurred in activations (Act (P)) destroy the task since all action steps will be greatly affected throughout the flight.
Weights are more sensitive to transient faults than activations (Act (T)) due to its massive reuse in convolutional layers. A fault in filter will be multiplied across whole feature map, therefore impacts larger value subsets.

\noindent \textbf{Different layer types.} The sensitivity of different positions and types of layers is evaluated in Fig.~\ref{fig:pedra_inference}d by injecting bit-flips into each layer individually. We find that the first two convolutional layers exhibit higher resiliency to transient faults. The reason is max-pooling and ReLU activation are implemented at the end of the first two layers, which can discard the faulty values and mask errors. However, there's no pooling layer after the third convolutional layer. Therefore, latter layers are generally more vulnerable to faults since faults have a relatively lower probability of being masked. The second fully-connected layer is the most vulnerable to bit-flips. This is because faults can spread to all neurons right after occurrence, and FC2 is the last layer that directly dictates the drone actions.

\noindent\textbf{Different data types.} Fig.~\ref{fig:pedra_inference}e demonstrates the fault tolerance of three fixed-point data types: Q (1,4,11), Q (1,7,8) and Q (1,10,5) (Q (sign, integer, fraction)) with the same flight quality baseline. We observe that Q (1,4,11) consistently exhibit higher resilience than Q (1,7,8) and Q (1,10,5). This is because Q (1,4,11) has a narrower range compared to others, and are already able to capture the range of weights without performance loss. Hence, a bit-flip in MSB part of Q (1,10,5) is likely to result in a larger deviation from zero and higher impacts. It is noted that to achieve higher resiliency, data types should be able to optimally capture the value range rather than pursuing an unnecessarily large range. This is in line with some recent works~\cite{tambe2019adaptivfloat,langroudi2020adaptive,tambe2020algorithm}.

\section{Fault Mitigation Techniques}
\label{sec:faullt_mitigatioin}
In consideration of the significant impact of faults, we propose and evaluate two fault mitigation schemes for training (Sec.~\ref{subsec: training_mitigation}) and inference (Sec.~\ref{subsec:inference_mitigation}), relying on our characterization observations.

\subsection{Training: Adaptive Exploration Rate Adjustment}
\label{subsec: training_mitigation}

\noindent\textbf{Detection.} Based on our observations in Sec.~\ref{sec:experiments}, transient faults can bring a sudden drop in reward and permanent faults can result in a continuous low reward. Therefore, we use the change in cumulative reward as an indicator of fault occurrence during training. 
For transient fault detection, if the cumulative reward drop exceeds $x$\% within $y$ consecutive episodes (where $x$ and $y$ are parameters), then it is supposed that fault occurs in the system. For permanent fault detection, after the exploration rate drops to a constant value and the agent goes into a stable greedy exploitation state, if the reward is still lower than 50\% of maximum value, then the permanent fault is supposed to be detected.

\noindent\textbf{Recovery.}
One of fundamental trade-offs impacting the RL training process is the exploration and exploitation ratio, which determines how agents explore the action space and exploit the knowledge at any given time step. Decaying-epsilon-greedy strategy is commonly used where the agent performs more explorations in the beginning and then explores less over time until a steady rate. This can make the policy converge fast and facilitate the learning process.

However, once faults occur, the agent needs more exploration actions to avoid being stuck in the wrong states and adapt itself to fault pattern. Therefore, we propose a dynamic exploration rate adjustment scheme to mitigate the negative fault impact during training. If transient faults are detected, the system will automatically increase its exploration rate (ER) followed by
\begin{align}
    ER_{new} = ER_{old} + \delta(ER)
    = ER_{old} +  \alpha \times min(f{(r)},f{(r)} f(t))
\end{align}
where $f{(r)}$ is normalized reward drop $f{(r)}=\Delta{r} /r_{max}$. $f{(t)}$ characterizes fault occurrence time $t$ defined as $f{(t)}=t/T$ ($T$ is episodes taken to reach steady exploitation in normal training). $ER_{new}$ is the adjusted exploration rate, and $\alpha$ is adjustment coefficient.

The agent needs more episodes to learn permanent fault patterns. Hence when permanent faults are detected, the agent will autonomously revert the exploration rate to initial and slow down its decreasing speed by $2^n \times$. This detection and recovery process will be triggered whenever the agent goes into the steady exploitation state, and $n$ is the number of detection times.

\noindent\textbf{Evaluation.} 
We measure the effectiveness of the proposed dynamic exploration rate adjustment technique by repeating experiments in Fig.~\ref{fig:gd_training_heatmap} where bit-flips are randomly injected with different BERs during training. We choose $x$=25, $y$=50 for detection, and $\alpha$=0.8 and $\alpha$=0.4 for tabular and NN-based methods respectively due to their self-healing ability difference shown in Sec.~\ref{subsec:gd}. The agent takes $T$=100 episodes to reach the steady exploitation state. 
\begin{figure}
\centering
\subfigure[Tabular-based approach]{
\begin{minipage}[b]{0.51\linewidth}
\includegraphics[width=\textwidth]{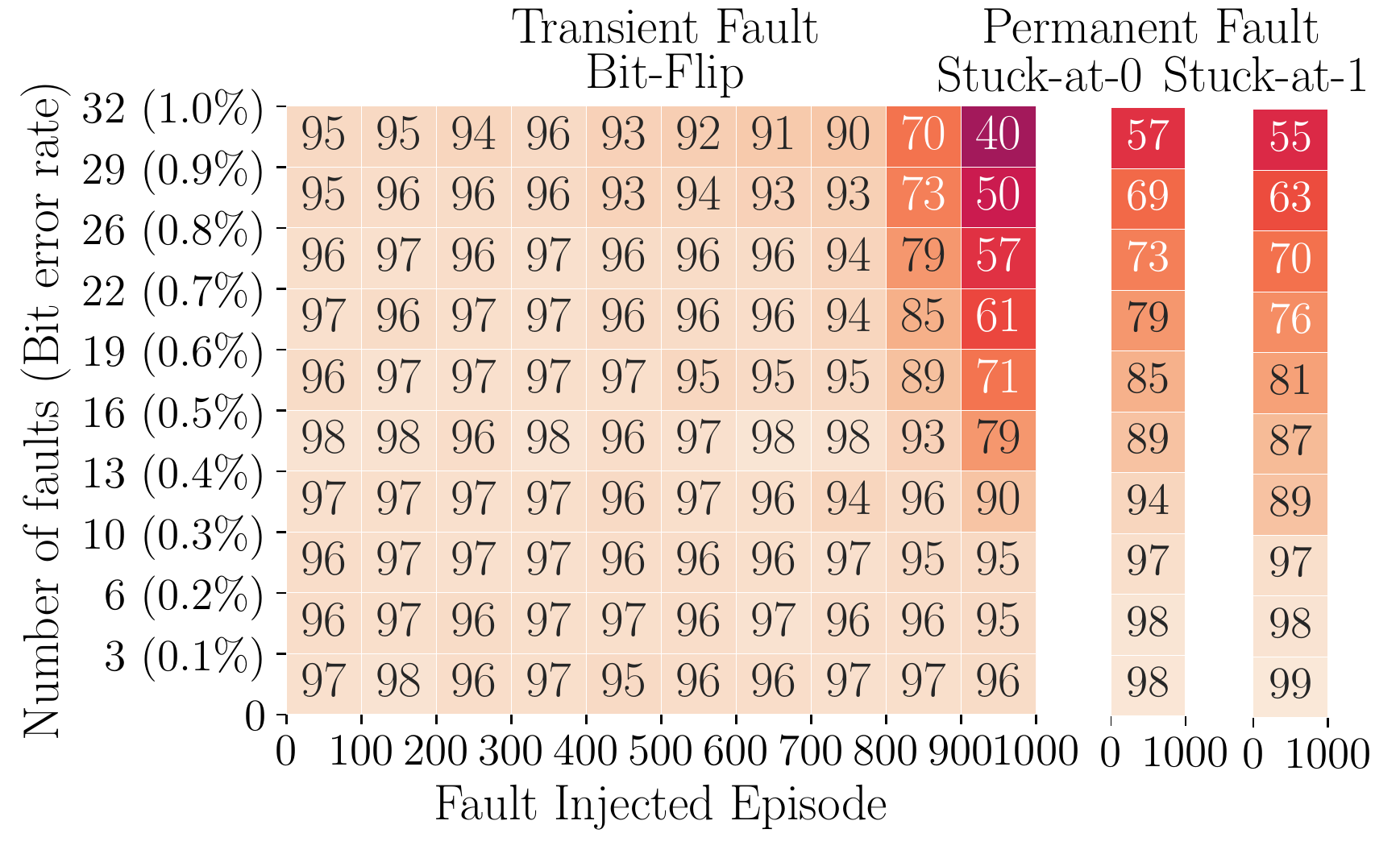}
\end{minipage}
}
\hspace{-0.115in}
\subfigure[NN-based approach]{
\begin{minipage}[b]{0.465\linewidth}
\includegraphics[width=\textwidth]{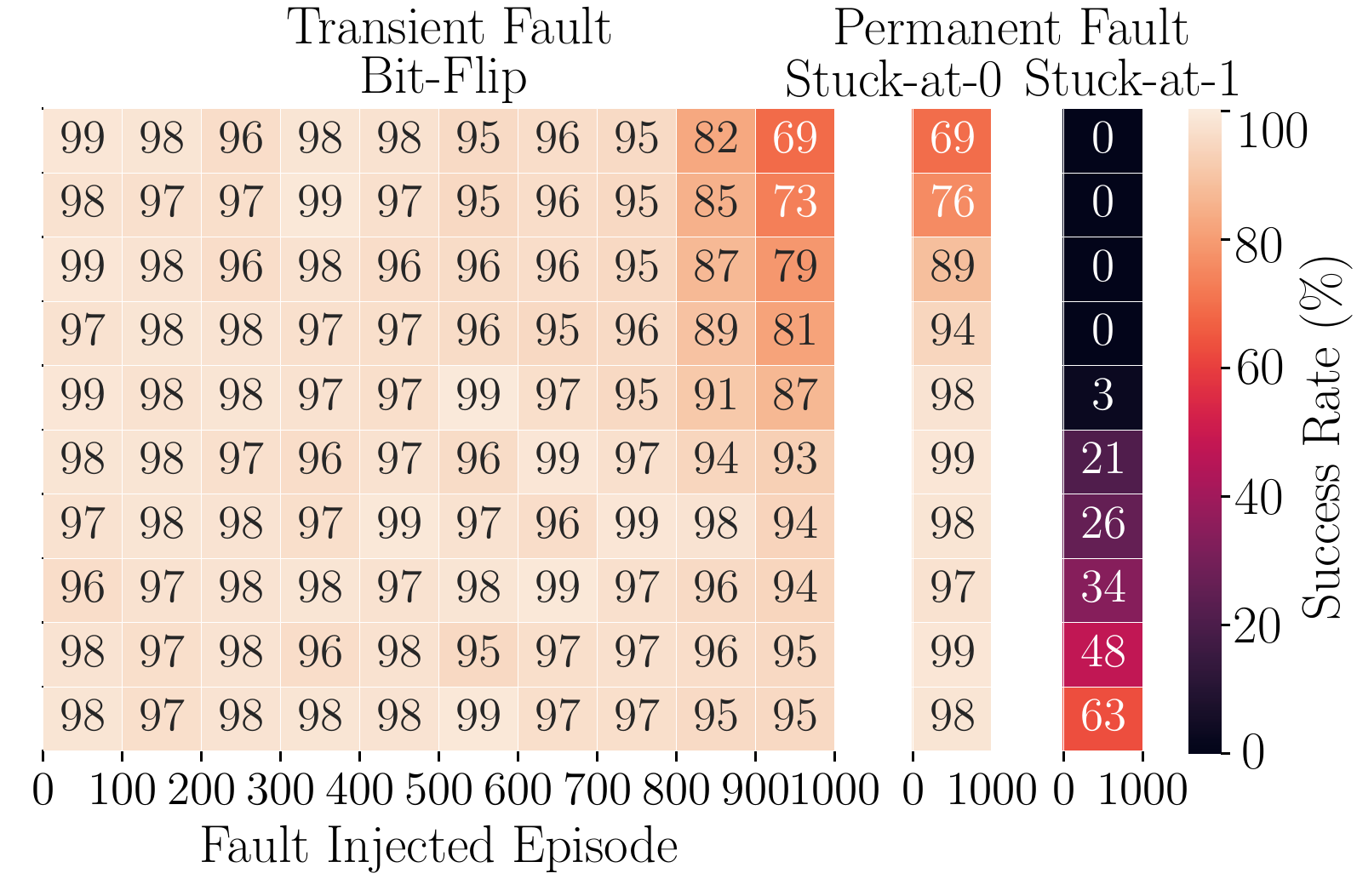}
\end{minipage}
}
\caption{The fault mitigation effect of dynamic exploration adjustment scheme on training.} \label{fig:mitigation_train}
\end{figure}

\begin{figure}
\centering
\subfigure[Tabular-based approach]{
\begin{minipage}[b]{0.34\linewidth}
\includegraphics[width=\textwidth]{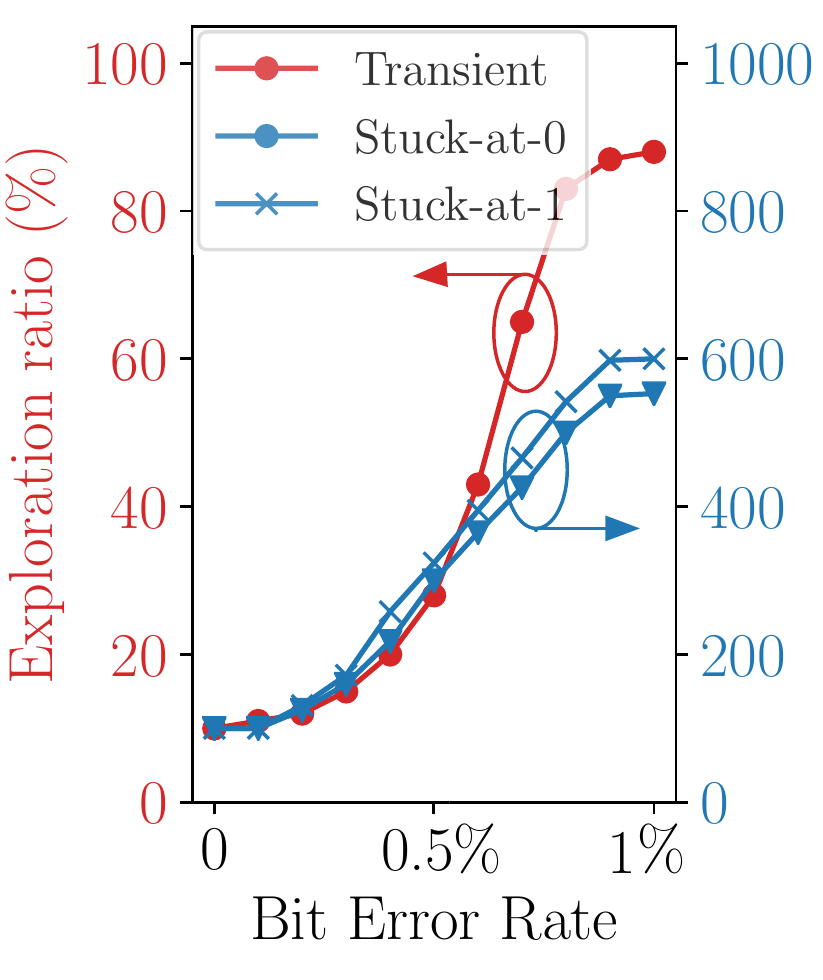}
\end{minipage}
}
\hspace{-0.13in}
\subfigure[NN-based approach]{
\begin{minipage}[b]{0.36\linewidth}
\includegraphics[width=\textwidth]{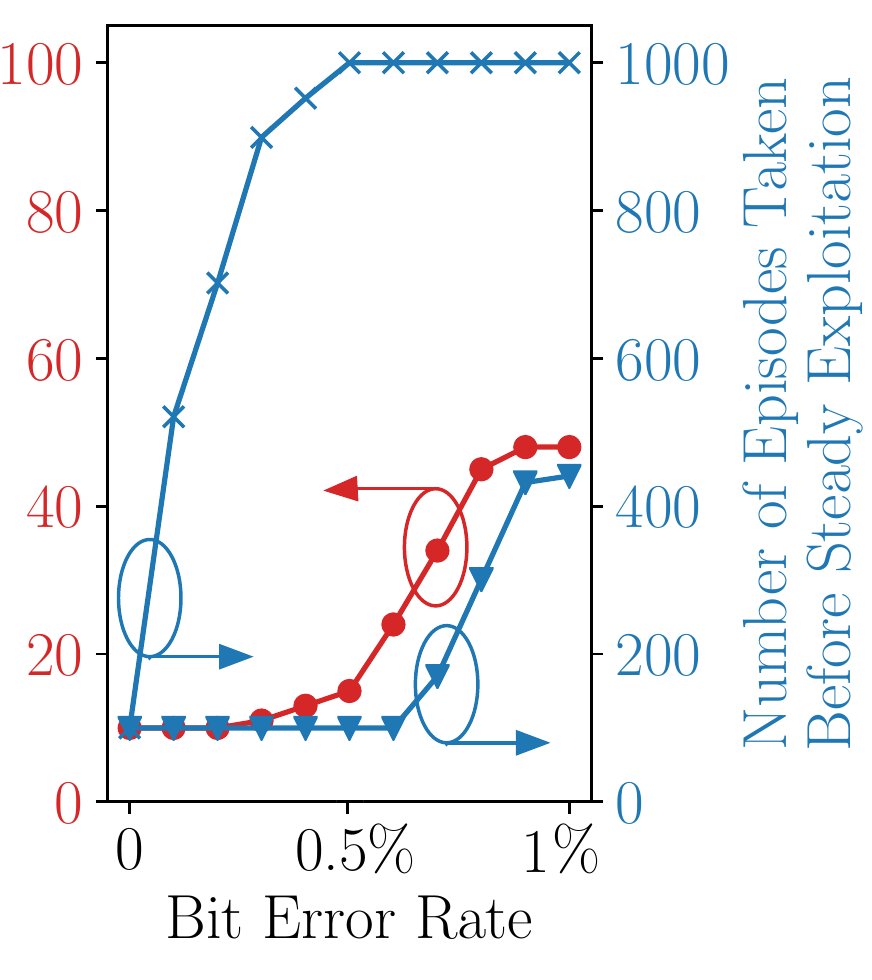}
\end{minipage}
}
\hspace{-0.13in}
\subfigure[Recovery Speed]{
\begin{minipage}[b]{0.285\linewidth}
\includegraphics[width=\textwidth]{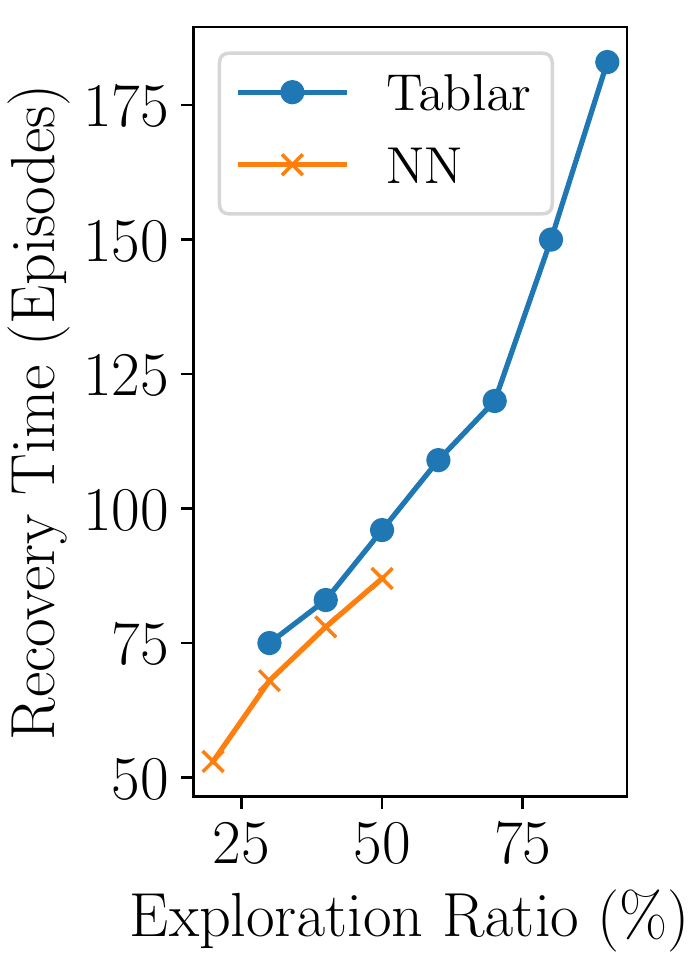}
\end{minipage}
}
\caption{The correlation between bit error rate, exploration ratio and transient fault recovery speed during training.} \label{fig:mitigation_train_correlation}
\vspace{0.1in}
\end{figure}

The evaluation results are shown in Fig.~\ref{fig:mitigation_train}. Compared with Fig.~\ref{fig:gd_training_heatmap}a and Fig.~\ref{fig:gd_training_heatmap}c, we find that almost all transient faults happened before 800 episodes become benign to the system, and the impact of faults happened in the last 200 episodes has also been greatly mitigated. The negative impact of permanent faults is also relieved by 10\%.

We also disclose the correlation between bit error rate, adjusted exploration rate and fault recovery speed. As shown in Fig.~\ref{fig:mitigation_train_correlation}a and Fig.~\ref{fig:mitigation_train_correlation}b, the agent automatically adjusts to a higher exploration rate when more transient faults detected (red), and adopts lower exploration decreasing speed when more permanent faults exist (blue). Note that if stuck-at-1 exists in NN-based system, the agent tends to conduct much more exploration to prevent itself stuck in the wrong states. Fig.~\ref{fig:mitigation_train_correlation}c demonstrates that agent usually needs more episodes to converge back if adjusting to a higher exploration rate. This is reasonable because if agent chooses low exploration rates under high BER scenarios, it may fail to recover from faults. The trade-off between recovery speed and general recovery success rate can be dynamically determined by agent in our proposed scheme.





\label{subsec:inference_mitigation}
\begin{figure}
\centering
\subfigure[Grid World problem]{
\begin{minipage}[b]{0.48\linewidth}
\includegraphics[width=\textwidth]{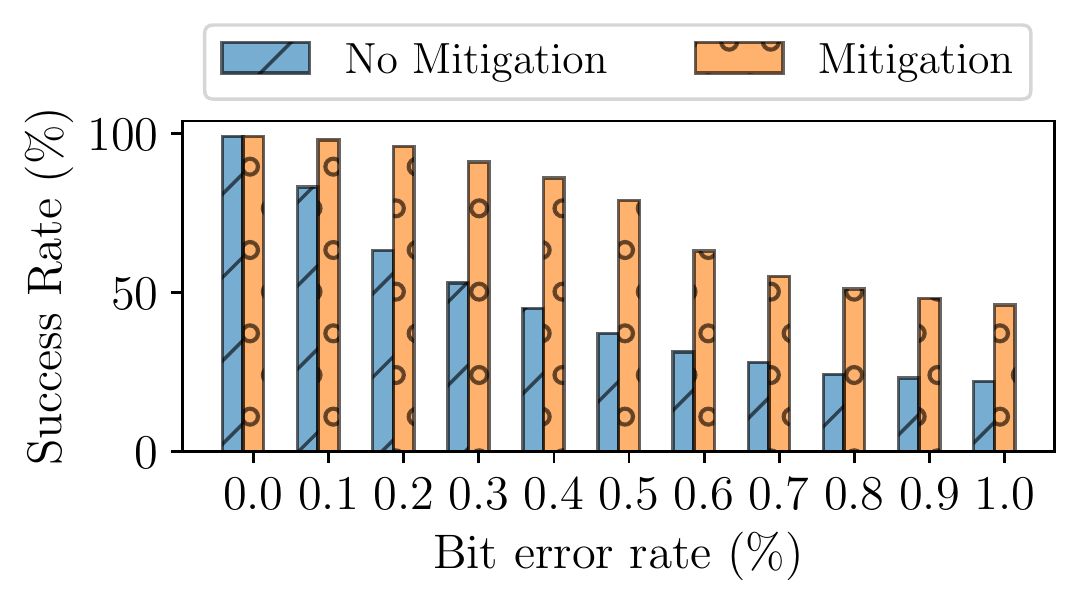}
\end{minipage}
}
\hspace{-0.1in}
\subfigure[Drone navigation problem]{
\begin{minipage}[b]{0.48\linewidth}
\includegraphics[width=\textwidth]{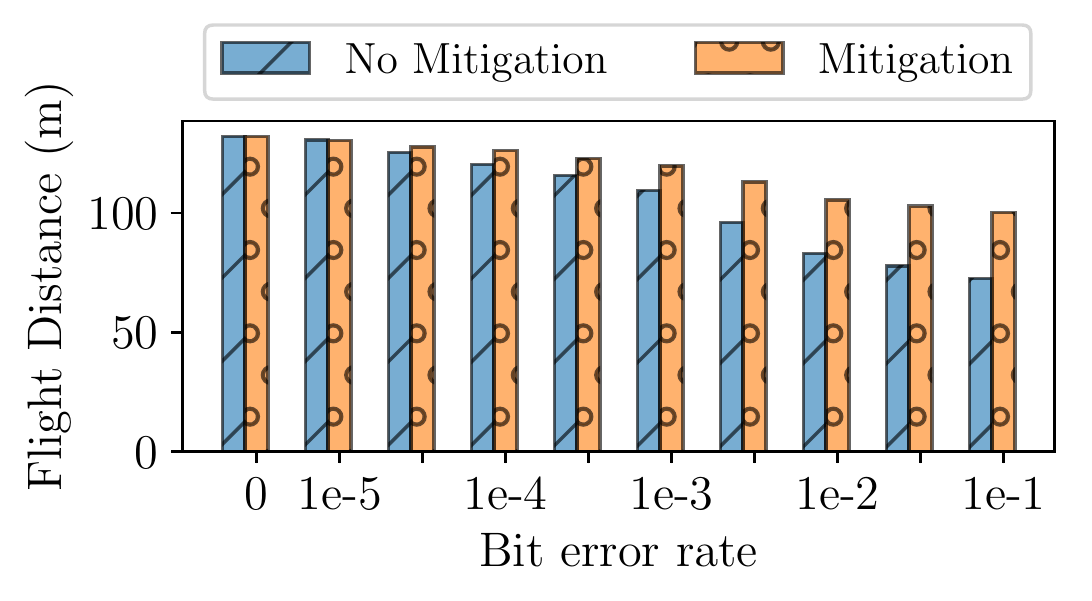}
\end{minipage}
}
\caption{The effectiveness of anomaly detection on inference.} \label{fig:infer_mitigation}
\vspace{0.1in}
\end{figure}

\subsection{Inference: Range-Based Anomaly Detection}
\noindent\textbf{Detection.} To detect unusual values, we adopt statistically anomaly detection technique that is widely used in safety-critical systems~\cite{rajbahadur2018survey}. Note that we perform value-level fault detection rather than conventional bit-level mismatch detection due to the observation that many bit-flips may eventually get masked or only result in small value deviations which won't impact final performance.

Based on the finding that faulty values with higher magnitude have a stronger destructive impact, we use the value range of parameters as an indicator to detect outliers. Once the policy is trained, the value ranges of each layer will be instrumented and derived (i.e. ($a_i, b_i$)), and 10\% detection margin will be applied on bound (i.e. ($1.1a_i, 1.1b_i$)) during inference. We only compare values composed of sign and integer bits based on the observation that fractional part has little impact, which can reduce the hardware cost further.

\noindent\textbf{Recovery.} Once any anomaly data is detected, an alarm signal will be raised, and the operations around this data will be skipped. The intuition is the sparsity nature of NN parameters, and the data with small magnitude has a higher possibility to become an outlier after bit-flip due to two's complement encoding method.

\noindent\textbf{Evaluation.} The effectiveness of our proposed anomaly detection scheme is evaluated in both Grid World and drone navigation problems, shown in Fig.~\ref{fig:infer_mitigation}. Transient faults are injected in NN weights. Compared to no mitigation cases, the agent's success rate increases by 2x in Grid World and drone's flight quality increases by 39\% after the fault mitigation is applied. Moreover, its runtime overhead is less than 3\% without requiring redundant bits as opposed to ECC.










\section{Conclusion}
\label{sec:conclusion}
This paper empirically investigates the end-to-end learning-based navigation system from the fault tolerance perspective, i.e., fault characterization and mitigation. We characterize the impact of transient and permanent faults on both training and inference stages through large-scale fault injection experiments on various training algorithms, data types and model structures. Based on the characterizations, we further propose two cost-effective fault mitigation techniques by dynamically adjusting the exploration rate during training and detecting anomaly ranges during inference. Our proposed method can improve system resilience by up to 2$\times$.


\section*{Acknowledgements}
We thank Muya Chang (from Georgia Tech) for his technical support.
This work was supported in part by C-BRIC and ADA, two of six centers in JUMP, a Semiconductor Research Corporation (SRC) program sponsored by DARPA.
%
\bibliographystyle{ieeetr}
\footnotesize{%
\bibliography{main}

\begin{thebibliography}{10}

\bibitem{liu2021robotic}
S.~Liu, Z.~Wan, B.~Yu, and Y.~Wang, ``Robotic computing on fpgas,'' {\em
  Synthesis Lectures on Computer Architecture}, vol.~16, no.~1, pp.~1--218,
  2021.

\bibitem{wan2021survey}
Z.~Wan, B.~Yu, T.~Y. Li, J.~Tang, Y.~Zhu, Y.~Wang, A.~Raychowdhury, and S.~Liu,
  ``A survey of fpga-based robotic computing,'' {\em IEEE Circuits and Systems
  Magazine}, vol.~21, no.~2, pp.~48--74, 2021.

\bibitem{suleiman2019navion}
A.~Suleiman, Z.~Zhang, L.~Carlone, S.~Karaman, and V.~Sze, ``Navion: A 2-mw
  fully integrated real-time visual-inertial odometry accelerator for
  autonomous navigation of nano drones,'' {\em IEEE Journal of Solid-State
  Circuits}, vol.~54, no.~4, pp.~1106--1119, 2019.

\bibitem{gao2021ielas}
T.~Gao, Z.~Wan, Y.~Zhang, B.~Yu, Y.~Zhang, S.~Liu, and A.~Raychowdhury,
  ``ielas: An elas-based energy-efficient accelerator for real-time stereo
  matching on fpga platform,'' in {\em 2021 IEEE 3rd International Conference
  on Artificial Intelligence Circuits and Systems (AICAS)}, pp.~1--4, IEEE,
  2021.

\bibitem{krishnan2021autosoc}
S.~Krishnan, T.~Tambe, Z.~Wan, and V.~J. Reddi, ``Autosoc: Automating
  algorithm-soc co-design for aerial robots,'' {\em arXiv preprint
  arXiv:2109.05683}, 2021.

\bibitem{wan2021energy}
Z.~Wan, Y.~Zhang, A.~Raychowdhury, B.~Yu, Y.~Zhang, and S.~Liu, ``An
  energy-efficient quad-camera visual system for autonomous machines on fpga
  platform,'' in {\em 2021 IEEE 3rd International Conference on Artificial
  Intelligence Circuits and Systems (AICAS)}, pp.~1--4, IEEE, 2021.

\bibitem{anwar2018navren}
M.~A. Anwar and A.~Raychowdhury, ``Navren-rl: Learning to fly in real
  environment via end-to-end deep reinforcement learning using monocular
  images,'' in {\em 2018 25th International Conference on Mechatronics and
  Machine Vision in Practice (M2VIP)}, pp.~1--6, IEEE, 2018.

\bibitem{krishnan2019air}
S.~Krishnan, B.~Borojerdian, W.~Fu, A.~Faust, and V.~J. Reddi, ``Air learning:
  An ai research platform for algorithm-hardware benchmarking of autonomous
  aerial robots,'' {\em arXiv preprint arXiv:1906.00421}, 2019.

\bibitem{duisterhof2019learning}
B.~P. Duisterhof, S.~Krishnan, J.~J. Cruz, C.~R. Banbury, W.~Fu, A.~Faust,
  G.~C. de~Croon, and V.~J. Reddi, ``Learning to seek: Autonomous source
  seeking with deep reinforcement learning onboard a nano drone
  microcontroller,'' {\em arXiv preprint arXiv:1909.11236}, 2019.

\bibitem{amaravati201855}
A.~Amaravati, S.~B. Nasir, J.~Ting, I.~Yoon, and A.~Raychowdhury, ``A 55-nm,
  1.0--0.4 v, 1.25-pj/mac time-domain mixed-signal neuromorphic accelerator
  with stochastic synapses for reinforcement learning in autonomous mobile
  robots,'' {\em IEEE Journal of Solid-State Circuits}, vol.~54, no.~1,
  pp.~75--87, 2018.

\bibitem{krishnan2021machine}
S.~Krishnan, Z.~Wan, K.~Bharadwaj, P.~Whatmough, A.~Faust, S.~Neuman, G.-Y.
  Wei, D.~Brooks, and V.~J. Reddi, ``Machine learning-based automated design
  space exploration for autonomous aerial robots,'' {\em arXiv preprint
  arXiv:2102.02988}, 2021.

\bibitem{anwar2020autonomous}
A.~Anwar and A.~Raychowdhury, ``Autonomous navigation via deep reinforcement
  learning for resource constraint edge nodes using transfer learning,'' {\em
  IEEE Access}, vol.~8, pp.~26549--26560, 2020.

\bibitem{tawada2015bit}
M.~Tawada, S.~Kimura, M.~Yanagisawa, and N.~Togawa, ``A bit-write reduction
  method based on error-correcting codes for non-volatile memories,'' in {\em
  The 20th Asia and South Pacific Design Automation Conference}, pp.~496--501,
  IEEE, 2015.

\bibitem{matsuo2018dual}
I.~B.~M. Matsuo, L.~Zhao, and W.-J. Lee, ``A dual modular redundancy scheme for
  cpu--fpga platform-based systems,'' {\em IEEE Transactions on Industry
  Applications}, vol.~54, no.~6, pp.~5621--5629, 2018.

\bibitem{bannon2019computer}
P.~Bannon, G.~Venkataramanan, D.~D. Sarma, and E.~Talpes, ``Computer and
  redundancy solution for the full self-driving computer,'' in {\em 2019 IEEE
  Hot Chips 31 Symposium (HCS)}, pp.~1--22, IEEE Computer Society, 2019.

\bibitem{toschi2019characterizing}
A.~Toschi, M.~Sanic, J.~Leng, Q.~Chen, C.~Wang, and M.~Guo, ``Characterizing
  perception module performance and robustness in production-scale autonomous
  driving system,'' in {\em IFIP International Conference on Network and
  Parallel Computing}, pp.~235--247, Springer, 2019.

\bibitem{jha2019ml}
S.~Jha, S.~Banerjee, T.~Tsai, S.~K. Hari, M.~B. Sullivan, Z.~T. Kalbarczyk,
  S.~W. Keckler, and R.~K. Iyer, ``Ml-based fault injection for autonomous
  vehicles: A case for bayesian fault injection,'' in {\em 2019 49th annual
  IEEE/IFIP international conference on dependable systems and networks (DSN)},
  pp.~112--124, IEEE, 2019.

\bibitem{hsiao2021mavfi}
Y.-S. Hsiao, Z.~Wan, T.~Jia, R.~Ghosal, A.~Raychowdhury, D.~Brooks, G.-Y. Wei,
  and V.~J. Reddi, ``Mavfi: An end-to-end fault analysis framework with anomaly
  detection and recovery for micro aerial vehicles,'' {\em arXiv preprint
  arXiv:2105.12882}, 2021.

\bibitem{reagen2018ares}
B.~Reagen, U.~Gupta, L.~Pentecost, P.~Whatmough, S.~K. Lee, N.~Mulholland,
  D.~Brooks, and G.-Y. Wei, ``Ares: A framework for quantifying the resilience
  of deep neural networks,'' in {\em 2018 55th ACM/ESDA/IEEE Design Automation
  Conference (DAC)}, pp.~1--6, IEEE, 2018.

\bibitem{mahmoud2020pytorchfi}
A.~Mahmoud, N.~Aggarwal, A.~Nobbe, J.~R.~S. Vicarte, S.~V. Adve, C.~W.
  Fletcher, I.~Frosio, and S.~K.~S. Hari, ``Pytorchfi: A runtime perturbation
  tool for dnns,'' in {\em 2020 50th Annual IEEE/IFIP International Conference
  on Dependable Systems and Networks Workshops (DSN-W)}, pp.~25--31, IEEE,
  2020.

\bibitem{chen2020tensorfi}
Z.~Chen, N.~Narayanan, B.~Fang, G.~Li, K.~Pattabiraman, and N.~DeBardeleben,
  ``Tensorfi: A flexible fault injection framework for tensorflow
  applications,'' in {\em 2020 IEEE 31st International Symposium on Software
  Reliability Engineering (ISSRE)}, pp.~426--435, IEEE, 2020.

\bibitem{li2017understanding}
G.~Li, S.~K.~S. Hari, M.~Sullivan, T.~Tsai, K.~Pattabiraman, J.~Emer, and S.~W.
  Keckler, ``Understanding error propagation in deep learning neural network
  (dnn) accelerators and applications,'' in {\em Proceedings of the
  International Conference for High Performance Computing, Networking, Storage
  and Analysis}, pp.~1--12, 2017.

\bibitem{hudson2018fault}
S.~Hudson, R.~S. Sundar, and S.~Koppu, ``Fault control using triple modular
  redundancy (tmr),'' in {\em Progress in Computing, Analytics and Networking},
  pp.~471--480, Springer, 2018.

\bibitem{torres2017fault}
C.~Torres-Huitzil and B.~Girau, ``Fault and error tolerance in neural networks:
  A review,'' {\em IEEE Access}, vol.~5, pp.~17322--17341, 2017.

\bibitem{salami2018resilience}
B.~Salami, O.~S. Unsal, and A.~C. Kestelman, ``On the resilience of rtl nn
  accelerators: Fault characterization and mitigation,'' in {\em 2018 30th
  International Symposium on Computer Architecture and High Performance
  Computing (SBAC-PAD)}, pp.~322--329, IEEE, 2018.

\bibitem{whatmough201714}
P.~N. Whatmough, S.~K. Lee, H.~Lee, S.~Rama, D.~Brooks, and G.-Y. Wei, ``A 28nm
  soc with a 1.2 ghz 568nj/prediction sparse deep-neural-network engine with>
  0.1 timing error rate tolerance for iot applications,'' in {\em 2017 IEEE
  International Solid-State Circuits Conference (ISSCC)}, pp.~242--243, IEEE,
  2017.

\bibitem{krishnan2019quantized}
S.~Krishnan, S.~Chitlangia, M.~Lam, Z.~Wan, A.~Faust, and V.~J. Reddi,
  ``Quantized reinforcement learning (quarl),'' {\em arXiv preprint
  arXiv:1910.01055}, 2019.

\bibitem{krishnan2020sky}
S.~Krishnan, Z.~Wan, K.~Bhardwaj, P.~Whatmough, A.~Faust, G.-Y. Wei, D.~Brooks,
  and V.~J. Reddi, ``The sky is not the limit: A visual performance model for
  cyber-physical co-design in autonomous machines,'' {\em IEEE Computer
  Architecture Letters}, vol.~19, no.~1, pp.~38--42, 2020.

\bibitem{tambe2019adaptivfloat}
T.~Tambe, E.-Y. Yang, Z.~Wan, Y.~Deng, V.~J. Reddi, A.~Rush, D.~Brooks, and
  G.-Y. Wei, ``Adaptivfloat: A floating-point based data type for resilient
  deep learning inference,'' {\em arXiv preprint arXiv:1909.13271}, 2019.

\bibitem{langroudi2020adaptive}
H.~F. Langroudi, V.~Karia, J.~L. Gustafson, and D.~Kudithipudi, ``Adaptive
  posit: Parameter aware numerical format for deep learning inference on the
  edge,'' in {\em Proceedings of the IEEE/CVF Conference on Computer Vision and
  Pattern Recognition Workshops}, pp.~726--727, 2020.

\bibitem{tambe2020algorithm}
T.~Tambe, E.-Y. Yang, Z.~Wan, Y.~Deng, V.~J. Reddi, A.~Rush, D.~Brooks, and
  G.-Y. Wei, ``Algorithm-hardware co-design of adaptive floating-point
  encodings for resilient deep learning inference,'' in {\em 2020 57th ACM/IEEE
  Design Automation Conference (DAC)}, pp.~1--6, IEEE, 2020.

\bibitem{rajbahadur2018survey}
G.~K. Rajbahadur, A.~J. Malton, A.~Walenstein, and A.~E. Hassan, ``A survey of
  anomaly detection for connected vehicle cybersecurity and safety,'' in {\em
  2018 IEEE Intelligent Vehicles Symposium (IV)}, pp.~421--426, IEEE, 2018.

\end{thebibliography}
}

\end{document}